# The Computational Complexity of Probabilistic Planning

**Michael L. Littman**                                    MLITTMAN@CS.DUKE.EDU
*Department of Computer Science, Duke University*
*Durham, NC 27708-0129 USA*

**Judy Goldsmith**                                        GOLDSMIT@CS.ENGR.UKY.EDU
*Department of Computer Science, University of Kentucky*
*Lexington, KY 40506-0046 USA*

**Martin Mundhenk**                                       MUNDHENK@TI.UNI-TRIER.DE
*FB4 - Theoretische Informatik, Universität Trier*
*D-54286 Trier, GERMANY*

## Abstract

We examine the computational complexity of testing and finding small plans in probabilistic planning domains with both flat and propositional representations. The complexity of plan evaluation and existence varies with the plan type sought; we examine totally ordered plans, acyclic plans, and looping plans, and partially ordered plans under three natural definitions of plan value. We show that problems of interest are complete for a variety of complexity classes: PL, P, NP, co-NP, PP, $\text{NP}^{\text{PP}}$, co-$\text{NP}^{\text{PP}}$, and PSPACE. In the process of proving that certain planning problems are complete for $\text{NP}^{\text{PP}}$, we introduce a new basic $\text{NP}^{\text{PP}}$-complete problem, E-MAJSAT, which generalizes the standard Boolean satisfiability problem to computations involving probabilistic quantities; our results suggest that the development of good heuristics for E-MAJSAT could be important for the creation of efficient algorithms for a wide variety of problems.

## 1. Introduction

Recent work in artificial-intelligence planning has addressed the problem of finding effective plans in domains in which operators have probabilistic effects (Drummond & Bresina, 1990; Mansell, 1993; Draper, Hanks, & Weld, 1994; Koenig & Simmons, 1994; Goldman & Boddy, 1994; Kushmerick, Hanks, & Weld, 1995; Boutilier, Dearden, & Goldszmidt, 1995; Dearden & Boutilier, 1997; Kaelbling, Littman, & Cassandra, 1998; Boutilier, Dean, & Hanks, 1998). Here, an "effective" or "successful" plan is one that reaches a goal state with sufficient probability. In *probabilistic propositional planning*, operators are specified in a Bayes network or an extended STRIPS-like notation, and the planner seeks a recipe for choosing operators to achieve a goal configuration with some user-specified probability. This problem is closely related to that of solving a Markov decision process (Puterman, 1994) when it is expressed in a compact representation.

In previous work (Goldsmith, Lusena, & Mundhenk, 1996; Littman, 1997a), we examined the complexity of determining whether an effective plan exists for completely observable domains; the problem is EXP-complete in its general form and PSPACE-complete when limited to polynomial-depth plans. (A polynomial-depth, or polynomial-horizon, plan is one that takes at most a polynomial number of actions before terminating.) For these results, plans are permitted to be arbitrarily large objects—there is no restriction that a valid plan need have any sort of compact (polynomial-size) representation.

Because they place no restrictions on the size of valid plans, these earlier results are not directly applicable to the problem of *finding* valid plans. It is possible, for example, that for a given planning domain, the only valid plans require exponential space (and exponential time) to write down. Knowing whether or not such plans exist is simply not very important because they are intractable to express.

In the present paper, we consider the complexity of a more practical and realistic problem—that of determining whether or not a plan exists in a given restricted form and of a given restricted size. The plans we consider take several possible forms that have been used in previous planning work: totally ordered plans, partially ordered plans, (totally ordered) conditional plans, and (totally order) looping plans. In all cases, we limit our attention to plans that can be expressed in size bounded by a polynomial in the size of the specification of the problem. This way, once we determine that a plan exists, we can use this information to try to write it down in a reasonable amount of time and space.

In the deterministic planning literature, several authors have addressed the computational complexity of determining whether a valid plan exists, of determining whether a plan exists of a given cost, and of finding the valid plans themselves under a variety of assumptions (Chapman, 1987; Bylander, 1994; Erol, Nau, & Subrahmanian, 1995; Bäckström, 1995; Bäckström & Nebel, 1995). These results provide lower bounds (hardness results) for analogous probabilistic planning problems since deterministic planning is a special case. In deterministic planning, optimal plans can be represented by a simple sequence of operators (a totally ordered plan). In probabilistic planning, a good conditional plan will often perform better than any totally ordered (unconditional) plan; therefore, we need to consider the complexity of the planning process for a richer set of plan structures.

For ease of discussion, we only explicitly describe the case of planning in completely observable domains. This means that the state of the world is known at all times during plan execution, in spite of the uncertainty of state transitions. We know that the state of the system is sufficient information for choosing actions optimally (Puterman, 1994), however, representing such a universal plan is often impractical in propositional domains in which the size of the state space is exponential in the size of the domain representation. For this reason, we consider other types of plan structures based on simple finite-state machines. Because the type of plans we consider do not necessarily use the full state of the system to make every decision, our results carry over to partially observable domains, although we do not explore this fact in detail in the present work.

The computational problems we look at are complete for a variety of complexity classes ranging from PL (probabilistic logspace) to PSPACE. Two results are deserving of special mention because they concern problems closely related to ones being actively addressed by artificial-intelligence researchers; first, the problem of evaluating a totally ordered plan in a compactly represented planning domain is PP-complete.[1] A compactly represented planning domain is one that is described by a two-stage temporal Bayes network (Boutilier et al., 1998) or similar notation.

---

1. The class PP is closely related to the somewhat more familiar #P; Toda (1991) showed that $P^{\#P} = P^{PP}$. Roughly speaking, this means that #P and PP are equally powerful when used as oracles. The counting class #P has already been recognized by the artificial-intelligence community as an important complexity class in computations involving probabilistic quantities, such as belief-network inference (Roth, 1996).

Second, the problem of determining whether a valid totally ordered plan exists for a compactly represented planning domain is NP$^{\text{PP}}$-complete. Whereas the class NP can be thought of as the set of problems solvable by guessing the answer and checking it in polynomial time, the class NP$^{\text{PP}}$ can be thought of as the set of problems solvable by guessing the answer and checking it using a probabilistic polynomial-time (PP) computation. It is likely that NP$^{\text{PP}}$ characterizes many problems of interest in the area of uncertainty in artificial intelligence; this paper and earlier work (Goldsmith et al., 1996; Mundhenk, Goldsmith, & Allender, 1997a; Mundhenk, Goldsmith, Lusena, & Allender, 1997b) give initial evidence of this.

## 1.1 Planning-Domain Representations

A probabilistic planning domain $M = \langle \mathcal{S}, s_0, \mathcal{A}, t, \mathcal{G} \rangle$ is characterized by a finite set of states $\mathcal{S}$, an initial state $s_0 \in \mathcal{S}$, a finite set of operators or actions $\mathcal{A}$, and a set of goal states $\mathcal{G} \subseteq \mathcal{S}$. The application of an action $a$ in a state $s$ results in a probabilistic transition to a new state $s'$ according to the probability transition function $t$, where $t(s, a, s')$ is the probability that state $s'$ is reached from state $s$ when action $a$ is taken. The objective is to choose actions, one after another, to move from the initial state $s_0$ to one of the goal states with probability above some threshold $\theta$.[2] The state of the system is known at all times (fully observable) and so can be used to choose the action to apply.

We are concerned with two main representations for planning domains: *flat* representations, which enumerate states explicitly, and *propositional* representations (sometimes called compact, structured, or factored representations), which view states as assignments to a set of Boolean state variables or propositions. Propositional representations can represent many domains exponentially more compactly than can flat representations.

In the flat representation, the transition function $t$ is represented by a collection of $|\mathcal{S}| \times |\mathcal{S}|$ matrices,[3] one for each action. In the propositional representation, this type of $|\mathcal{S}| \times |\mathcal{S}|$ matrix would be huge, so the transition function must be expressed another way. In the probabilistic planning literature, two popular representations for propositional planning domains are probabilistic state-space operators (PSOs) (Kushmerick et al., 1995) and two-stage temporal Bayes networks (2TBNs) (Boutilier et al., 1995). Although these representations differ in the type of planning domains they can express naturally (Boutilier et al., 1998), they are computationally equivalent; a planning domain expressed in one representation can be converted in polynomial time to an equivalent planning domain expressed in the other with at most a polynomial increase in representation size (Littman, 1997a).

In this work, we focus on a propositional representation called the sequential-effects-tree representation (ST) (Littman, 1997a), which is a syntactic variant of 2TBNs with conditional probability tables represented as trees (Boutilier et al., 1995, 1998). This representation is equivalent to 2TBNs and PSOs and simplifies the presentation of our results.

---

2. It is also possible to formulate the objective as one of maximizing expected total discounted reward (Boutilier et al., 1995), but the two formulations are essentially polynomially equivalent (Condon, 1992). The only difficulty is that compactly represented domains may require discount factors exponentially close to one for this equivalence to hold. This is discussed further in Section 5.

3. We assume that the number of bits used to represent the individual probability values isn't too large.

In ST, the effect of each action on each proposition is represented as a separate decision tree. For a given action $a$, the set of decision trees for the different propositions is ordered, so the decision tree for one proposition can refer to both the new and old values of previous propositions; this allows ST to represent any probability distribution. The leaves of a decision tree describe how the associated proposition changes as a function of the state and action, perhaps probabilistically. Section 1.2 gives a simple example of this representation.

As in other propositional representations, the states in the set of goal states $\mathcal{G}$ are not explicitly enumerated in ST. Instead, we define a *goal set*, which is a set of propositions such that any state in which all the goal-set propositions are true is considered a goal state. The set of actions $\mathcal{A}$ is explicitly enumerated in ST, just as it is in the flat representation.

The ST representation of a planning domain $M = \langle \mathcal{S}, s_0, \mathcal{A}, t, \mathcal{G} \rangle$ can be defined more formally as $\mathbb{M} = \langle \mathbb{P}, \mathbb{I}, \mathcal{A}, \mathbb{T}, \mathbb{G} \rangle$ (we use blackboard-bold font to stand for an ST representation on a domain). Here, $\mathbb{P}$ is a finite set of distinct propositions. The set of states $\mathcal{S}$ is the power set of $\mathbb{P}$; the propositions in $s \in \mathcal{S}$ are said to be "true" in $s$. The set $\mathbb{I} \subseteq \mathbb{P}$ is the initial state. The set $\mathbb{G}$ is the goal set, so the set of goal states $\mathcal{G}$ is the set of states $s$ such that $\mathbb{G} \subseteq s$.

The transition function $t$ is represented by a function $\mathbb{T}$, which maps each action in $\mathcal{A}$ to an ordered sequence of $|\mathbb{P}|$ binary decision trees. Each of these decision trees has a distinct label proposition, decision propositions at the nodes (optionally labeled with the suffix "**:new**"), and probabilities at the leaves. The $i$th decision tree $\mathbb{T}(a)_i$ for action $a$ defines the transition probabilities $t(s, a, s')$ as follows. For the $i$th decision tree, let $\mathbf{p}_i$ be its label proposition. Define $\rho_i$ to be the value of the leaf node found by traversing decision tree $\mathbb{T}(a)_i$, taking the left branch if the decision proposition is in $s$ (or $s'$ if the decision proposition has the "**:new**" suffix) and the right branch otherwise. Finally, we let

$$t(s, a, s') = \prod_i \left\{ \begin{array}{ll} \rho_i, & \text{if } \mathbf{p}_i \in s', \\ 1 - \rho_i, & \text{otherwise.} \end{array} \right. \tag{1}$$

This definition of $t$ constitutes a well-defined probability distribution over $s'$ for each $a$ and $s$.

To insure the validity of the representation, we only allow "**p:new**" to appear as a decision proposition in $\mathbb{T}(a)_i$ if $\mathbf{p}$ is the label proposition for some decision tree $\mathbb{T}(a)_j$ for $j < i$. For this reason, the order of the decision trees in $\mathbb{T}(a)$ is significant. To put this another way, a proposition only has a new value after this new value has been defined by some decision tree.

The complexity results we derive for ST apply also to PSOs, 2TBNs, and all other computationally equivalent representations. They also hold for the "succinct representation," a propositional representation popular in the complexity-theory literature, which captures the set of transition matrices as a function, most commonly represented by a Boolean circuit that computes that function. ST can straightforwardly be represented as a Boolean circuit, and, in the proof of Theorem 6, we show how to represent particular Boolean circuits in ST. Thus, although we have not shown that the succinct representation is formally equivalent to ST, the two representations are closely related; the proofs we give for ST need to be changed only slightly to work for the succinct representation (Goldsmith, Littman, & Mundhenk, 1997a, 1997b; Mundhenk et al., 1997b). Our results require that we restrict the succinct representation to generate transition probabilities with at most a polynomial number of bits; the results may be different for other circuit-based representations that can represent probabilities with an exponential number of bits (Mundhenk et al., 1997a).

## 1.2 Example Domain

To help make these domain-representation ideas more concrete, we present the following simple probabilistic planning domain based on the problem of building a sand castle at the beach. There are a total of four states in the domain, described by combinations of two Boolean propositions, **moat** and **castle** (propositions appear in boldface). The proposition **moat** signifies that a moat has been dug in the sand, and the proposition **castle** signifies that the castle has been built. In the initial state, both **moat** and **castle** are false, and the goal set is {**castle**}.

There are two actions: dig-moat and erect-castle (actions appear in sans serif). Figure 1 illustrates these actions in ST. Executing dig-moat when **moat** is false causes **moat** to become true with probability 1/2; if **moat** is already true, dig-moat leaves it unchanged. The **castle** proposition in not affected. The dig-moat action is depicted in the left half of Figure 1.

The second action is erect-castle, which appears in the right half of Figure 1. The decision trees are numbered to allow sequential dependencies between their effects to be expressed. The first decision tree is for **castle**, which does not change value if it is already true when erect-castle is executed. Otherwise, the probability that it becomes true is dependent on whether **moat** is true; the castle is built with probability 1/2 if **moat** is true and only probability 1/4 if it is not. The idea here is that building a moat first protects the castle from being destroyed prematurely by the ocean waves.

The second decision tree is for the proposition **moat**. Because erect-castle cannot make **moat** become true, there is no effect when **moat** is false. On the other hand, if the moat exists, it may collapse as a result of trying to erect the castle. The label **castle:new** in the diagram refers to the value of the **castle** proposition *after* the first decision tree is evaluated. If the castle was already built when erect-castle was selected, the moat remains built with probability 3/4. If the castle had not been built, but erect-castle successfully builds it, **moat** remains true. Finally, if erect-castle fails to make **castle** true, **moat** becomes false with probability 1/2 and everything is destroyed.

Note that given an ST representation of a domain, we can perform a number of useful operations efficiently. First, given a state $s$ and action $a$, we can generate a next state $s'$ with the proper probabilities. This is accomplished by calculating the value of the propositions of $s'$ one at a time in the order given in the representation of $a$, flipping coins with the probabilities given in the leaves of the decision trees. Second, given a state $s$, action $a$, and state $s'$, we can compute $t(s, a, s')$, the probability that state $s'$ is reached from state $s$ when action $a$ is taken, via Equation 1.

## 1.3 Plan Types and Representations

We consider four classes of plans for probabilistic domains. *Totally ordered plans* are the most basic type, being a finite sequence of actions that must be executed in order; this type of plan ignores the state of the system. *Acyclic plans* generalize totally ordered plans to include conditional execution of actions. *Partially ordered plans* are a different way of generalizing totally ordered plans in which the precise sequence is left flexible (McAllester & Rosenblitt, 1991). *Looping plans* generalize acyclic plans to the case in which plan steps can be repeated (Smith & Williamson, 1995; Lin & Dean, 1995). This type of plan is also referred to as a plan graph or policy graph (Kaelbling et al., 1998).

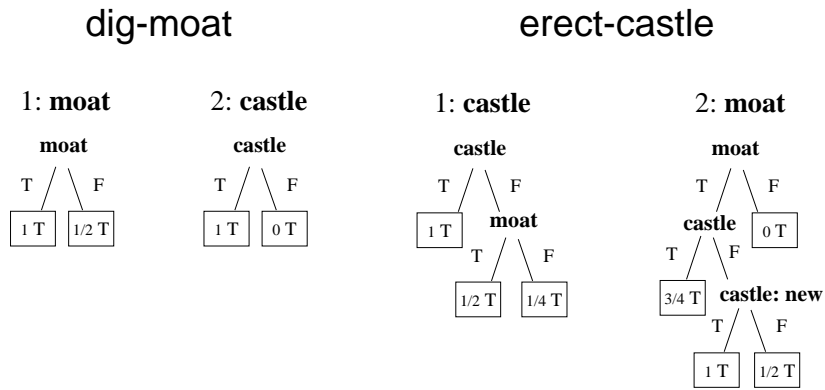

Figure 1: Sequential-effects-tree (ST) representation for the sand-castle domain

In the following sections, we prove computational complexity results concerning each of these plan types. The remainder of this section provides formal definitions of the plan types, illustrated in Figure 2 with examples for the sand-castle domain.

In its most general form, a plan (or policy, controller or transducer) is a program that outputs actions and takes as input information about the outcome of these actions. In this work, we consider only a particularly restricted finite-state-controller-based plan representation.

A *plan* $P$ for a planning domain $M = \langle \mathcal{S}, s_0, \mathcal{A}, t, \mathcal{G} \rangle$ can be represented by a structure $(V, v_0, E, \pi, \delta)$ consisting of a directed (multi) graph $(V, E)$ with initial node $v_0 \in V$, a labeling $\pi : V \to \mathcal{A}$ of plan nodes—called *plan steps*—to domain actions, and a labeling of edges with state sets $\delta : E \to \mathcal{P}(\mathcal{S})$ such that for every $v \in V$ with outgoing edges, $\bigcup_{v' \in V : (v,v') \in E} \delta(v, v') = \mathcal{S}$ and $\delta(v, v_1) \cap \delta(v, v_2) = \emptyset$ for all $v_1, v_2 \in V$, $v_1 \neq v_2$. Some plan steps have no outgoing edges at all—these are the *terminal steps*. Actions for terminal steps are not executed. Note that the function $\delta$ can be represented in a direct manner for flat domains, but for propositional domains, a more compact representation is needed. We assume that for propositional domains, edge labels are given as conjunctions of literals.

The behavior of plan $P$ in domain $M$ is as follows. The initial time step is $t = 0$. At time step $t \geq 0$, the domain is in state $s_t$ and the plan is at step $v_t$ ($s_0$ is defined by the planning domain, $v_0$ by the plan). Action $\pi(v_t)$ is executed, resulting in a transition to domain state $s_{t+1}$ with probability $t(s_t, \pi(v_t), s_{t+1})$. Plan step $v_{t+1}$ is chosen so that $s_{t+1} \in \delta(v_t, v_{t+1})$; the function $\delta$ tells the plan where to "go" next. At this point, the time-step index $t$ is incremented and the process repeats. This continues until a terminal step is reached in the plan.

One can understand the behavior of domain $M$ under plan $P$ in several different ways. The possible sequences of states of $M$ can be viewed as a tree: each node of the tree at depth $t$ is a state reachable from the initial state at time step $t$. Alternatively, one can view the state of $M$ at time step $t$ under plan $P$ as a probability distribution over $\mathcal{S}$. At time step 0, with probability 1 the process is in state $s_0$. The probability that $M$ is in state $s'$ at time step $t + 1$, $\Pr(s', t + 1)$, is the sum of the probabilities of all length $t + 1$ paths from $s_0$ to $s'$, i.e.,

$$\sum_{s_0, s_1, s_2, \ldots, s_t, s_{t+1} = s'} \prod_{j=1}^{t} t(s_j, a_j, s_{j+1}),$$

where $a_j$ is the action selected by plan $P$ at time $j$ given the observed sequence of state transitions $s_0, \ldots, s_j$. This view is useful in some of the later proofs.

Next, we formalize the probability that domain $M$ reaches a goal state under plan $P$. We need to introduce several notions. A "legal" sequence of states and steps applied is called a *trajectory*, i.e., for $M$ and $P$ this is a sequence $\alpha = \langle (s_i, v_i) \rangle_{i=0}^{k}$ of pairs with

- $t(s_i, \pi(v_i), s_{i+1}) > 0$ for $0 \leq i \leq k - 1$,

- $s_{i+1} \in \delta(v_i, v_{i+1})$ for $0 \leq i \leq k - 1$, and

- $v_0, \ldots, v_{k-1}$ are not terminal steps.

A *goal trajectory* is a trajectory that ends in a goal state of $M$, $s_k \in \mathcal{G}$. Note that each goal trajectory is finite. Thus, we can calculate the probability of a goal trajectory $\alpha = \langle (s_i, v_i) \rangle_{i=0}^{k}$ as $\Pr(\alpha) = \prod_{i=0}^{k-1} t(s_i, \pi(v_i), s_{i+1})$, given that $s_k \in \mathcal{G}$. The probability that $M$ reaches a goal state under plan $P$ is the sum of the probabilities of goal trajectories for $M$,

$$\Pr(M \text{ reaches a goal state under } P) := \sum_{\alpha \text{ goal trajectory}} \Pr(\alpha);$$

we call this the *value* of the plan.

We characterize a plan $P = (V, v_0, E, \pi, \delta)$ on the basis of the size and structure of its underlying graph $(V, E)$. If the graph $(V, E)$ contains no cycles, we call it an *acyclic plan*, otherwise it is a *looping plan*. It follows that an acyclic plan has a terminal step, and that a terminal step will be reached after no more than $|V|$ actions are taken; such plans can only be used for finite-horizon control. A *totally ordered plan* (sometimes called a "linear plan" or a "straight line" plan) is an acyclic plan with no more than one outgoing edge for each node in $V$. Such a plan is a simple path.

In this work, we also consider *partially ordered plans* (sometimes called "nonlinear" plans) that express an entire family of totally ordered plans. In this representation, the steps of the plan are given as a partial order (specified, for example, as a directed acyclic graph). This partial order represents a *set* of totally ordered plans: all totally ordered sequences of plan steps consistent with the partial order that consist of *all* steps of the partially ordered plan. Each of these totally ordered plans has a value, and these values need not all be the same. As such, we have a choice in defining the value for a partially ordered plan. In this work, we consider the optimistic, pessimistic, and average interpretations. Let $\Omega(P)$ be the set of totally ordered sequences consistent with partial order plan $P$. Under the *optimistic* interpretation,

$$\text{The value of } P := \max_{p \in \Omega(P)} \Pr(M \text{ reaches a goal state under } p).$$

Under the *pessimistic* interpretation,

$$\text{The value of } P := \min_{p \in \Omega(P)} \text{Pr}(M \text{ reaches a goal state under } p).$$

Under the *average* interpretation,

$$\text{The value of } P := \frac{1}{|\Omega(P)|} \sum_{p \in \Omega(P)} \text{Pr}(M \text{ reaches a goal state under } p).$$

To illustrate these notions, Figure 2 gives plans of each type for the sand-castle domain described earlier. Initial nodes are marked an incoming arrow, and terminal steps are represented as filled circles. The 3-step totally ordered plan in Figure 2(a) successfully builds a sand castle with probability 0.4375. An acyclic plan is given in Figure 2(b), which succeeds with probability 0.46875 and executes dig-moat an average of 1.75 times. Note that it succeeds more often and with fewer actions on average than the totally ordered plan in Figure 2(a).

Figure 2(c) illustrates a partially ordered plan for the sand-castle domain. While this plan bears a superficial resemblance to the acyclic plan in Figure 2(b), it has a different interpretation. In particular, the plan in Figure 2(c) represents a set of totally ordered plans with five (non-terminal) plan steps (3 dig-moat steps and 2 erect-castle steps). In contrast to the solid arrows in Figure 2(b), which indicate flow of control, the dashed arrows in Figure 2(c) represent ordering constraints: each erect-castle step must be preceded by at least two dig-moat steps, for example.

Although there are $\binom{5}{2} = 10$ distinct ways of arranging the five plan steps in Figure 2(c) into a totally ordered plan, only two distinct totally ordered plans are consistent with the ordering constraints:

$$\text{dig-moat} \rightarrow \text{dig-moat} \rightarrow \text{dig-moat} \rightarrow \text{erect-castle} \rightarrow \text{erect-castle} \rightarrow \bullet$$

(success probability 0.65625) and

$$\text{dig-moat} \rightarrow \text{dig-moat} \rightarrow \text{erect-castle} \rightarrow \text{dig-moat} \rightarrow \text{erect-castle} \rightarrow \bullet$$

(success probability 0.671875). Thus, the optimistic success probability of this partially ordered plan is 0.671875, the pessimistic 0.65625. Note that the pessimistic interpretation is closely related to the standard interpretation in deterministic partial order planning (McAllester & Rosenblitt, 1991), in which a partially ordered plan is considered successful only if *all* its consistent totally ordered plans are successful. The average success probability is 0.6614583, here, because there are 4 orderings that yield the poorer plan described above, and 2 that yield the better one.

The looping plan in Figure 2(d) does not terminate until it succeeds in building a sand castle, which it will do with probability 1.0 eventually. Of course, not all looping plans succeed with probability 1; the totally ordered plan in Figure 2(a) and the acyclic plan in Figure 2(b) are special cases of such looping plans, for instance.

We define $|P|$ the *size of a plan P* to be the number of steps it contains. We define $|M|$ the *size of a domain M* to be the sum of the number of actions and states for a flat domain and the sum of the sizes of the ST decision trees for a propositional domain.

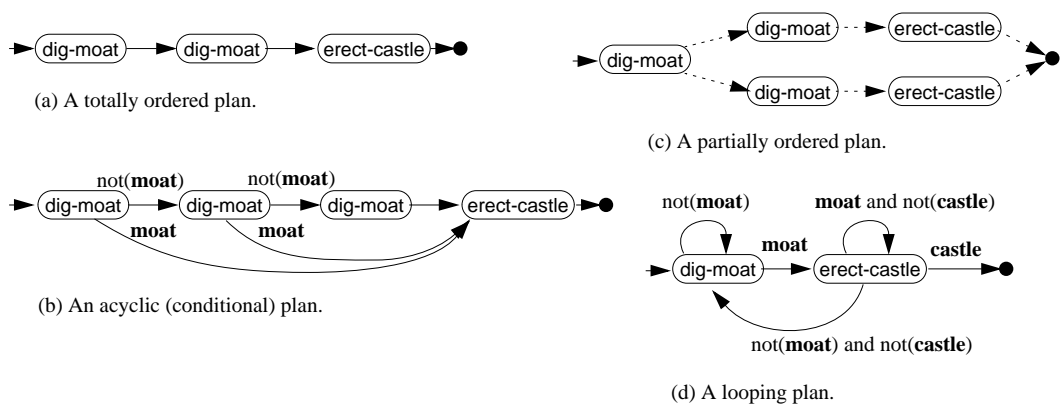

Figure 2: Example plans for the sand-castle domain

We consider the following decision problems. The *plan-evaluation problem* asks, given a domain $M$, a plan $P$ of size $|P| \leq |M|$, and threshold $\theta$, whether its value is greater than $\theta$, i.e., whether

$$\Pr(M \text{ reaches a goal state under } P) > \theta.$$

Note that the condition that $|P| \leq |M|$ is just a technical one—we simply want to use $|M|$ to represent the size of the problem. Given an instance in which $|P|$ is larger than $|M|$, we simply imagine "padding out" $|M|$ to make it larger. The important thing is that we are considering plans that are roughly the size of the description of the domain, and not the size of the number of states (which might be considerably larger).

The *plan-existence problem* asks, given domain $M$, threshold $\theta$, and size bound $z \leq |M|$, whether there exists a plan $P$ of size $z$ with value greater than $\theta$. Note that because we bound the size of the target plan, the complexity of plan generation is no more than that of plan existence; the technique of self-reduction can be used to construct a valid plan using polynomially many calls to an oracle for the decision problem.

Each of these decision problems has a different version for each type of domain (flat and propositional) and each type of plan category (looping, acyclic, totally ordered, and partially ordered under each of the three interpretations). We address all of these problems in the succeeding sections.

## 1.4 Complexity Classes

For definitions of complexity classes, reductions, and standard results from complexity theory, we refer the reader to Papadimitriou (1994).

Briefly, we are looking only at the complexity of decision problems (those with yes/no answers). The class P consists of problems that can be decided in polynomial time; that is, given an instance of the problem, there is a program for deciding whether the answer is yes or no that runs in polynomial time. The class NP contains the problems with polynomial-time checkable polynomial-size certificates: for any given instance and certificate, it can be checked in time polynomial in the size of the instance whether the certificate proves that the instance is in the NP set. This means that, if the answer to the instance is "yes," this can be shown in polynomial time given the right key. The class co-NP is the opposite—if the answer is "no," this can be shown in polynomial time given the right key.

A problem $X$ is $\mathcal{C}$-hard for some complexity class $\mathcal{C}$ if every problem in $\mathcal{C}$ can be reduced to it; to put it another way, a fast algorithm for $X$ can be used as a subroutine to solve any problem in $\mathcal{C}$ quickly. A problem is $\mathcal{C}$-complete if it is both $\mathcal{C}$-hard and in $\mathcal{C}$; these are the hardest problems in the class.

In the interest of being complete, we next give more detailed descriptions of the less familiar probabilistic and counting complexity classes we use in this work.

The class #L (Àlvarez & Jenner, 1993) is the class of functions $f$ such that, for some nondeterministic logarithmically space-bounded machine $N$, the number of accepting paths of $N$ on $x$ equals $f(x)$. The class #P is defined analogously as the class of functions $f$ such that, for some nondeterministic *polynomial-time*-bounded machine $N$, the number of accepting paths of $N$ on $x$ equals $f(x)$. Typical complete problems are computing the determinant for #L and computing the permanent for #P.

A function $f$ is defined to be in GapL if it is the *difference* $f = g - h$ of #L functions $g$ and $h$. While #L functions have nonnegative integer values by definition, GapL functions may have negative integer values (for example, if $g$ always returns zero).

Probabilistic logspace (Gill, 1977), PL, is the class of sets $A$ for which there exists a nondeterministic logarithmically space-bounded machine $N$ such that $x \in A$ if and only if the number of accepting paths of $N$ on $x$ is greater than its number of rejecting paths. In the original definition of PL, there is no time bound on computations; Borodin, Cook, and Pippenger (1983) later showed PL $\subseteq$ P. Jung (1985) proved that any set computable in probabilistic logspace is computable in probabilistic logspace where the PL machine has a simultaneous polynomial-time bound. In apparent contrast to P-complete sets, sets in PL are decidable using very fast parallel computations (Borodin et al., 1983).

Probabilistic polynomial time, PP, is defined analogously. A classic PP-complete problem is MAJSAT: given a Boolean formula in conjunctive normal form (CNF), does the majority of assignments satisfy it? According to Balcázar, Díaz, and Gabarró (1990), the PP-completeness of MAJSAT was shown in a combination of results from Gill (1977) and Simon (1975).

For polynomial-space-bounded computations, PSPACE equals probabilistic PSPACE, and #PSPACE is the same as the class of polynomial-space-computable functions (Ladner, 1989).

Note that L, NL, #L, PL and GapL are to logarithmic space what P, NP, #P, PP, and GapP are to polynomial time. Also, the notion of completeness we use in this paper relies on many-one reductions. In the case of PL, the reduction functions are logarithmic space; in the case of NP and above, they are polynomial time.

For any complexity classes $\mathcal{C}$ and $\mathcal{C}'$ the class $\mathcal{C}^{\mathcal{C}'}$ consists of those sets that are $\mathcal{C}$-*Turing reducible* to sets in $\mathcal{C}'$, i.e., sets that can be accepted with resource bounds specified by $\mathcal{C}$, using some problem in $\mathcal{C}'$ as a subroutine (oracle) with instantaneous output. For any class $\mathcal{C} \subseteq$ PSPACE, it is the case that NP$^{\mathcal{C}} \subseteq$ PSPACE, and therefore NP$^{\text{PSPACE}} =$ PSPACE.

The primary oracle-defined class we consider is NP$^{\text{PP}}$. It equals the "$\leq_m^{\text{NP}}$" closure of PP (Torán, 1991), which can be seen as the closure of PP under polynomial-time disjunctive reducibility with an exponential number of queries (each of the queries computable in polynomial time from its index in the list of queries). To simplify our completeness results for this class, we introduce a decision problem we call E-Majsat ("exists" Majsat), which generalizes the standard NP-complete satisfiability problem and the PP-complete Majsat. An E-Majsat instance is defined by a CNF Boolean formula $\phi$ on $n$ Boolean variables $x_1, \ldots, x_n$ and a number $k$ between 1 and $n$. The task is to decide whether there is an initial partial assignment to variables $x_1, \ldots, x_k$ so that the majority of assignments that extend that partial assignment satisfies $\phi$. We prove that this problem is $\mathrm{NP^{PP}}$-complete in the Appendix.

The complexity classes we consider satisfy the following containment properties and relations to other well-known classes:

$$\mathrm{L} \subseteq \mathrm{NL} \subseteq \mathrm{PL} \subseteq \mathrm{P} \subseteq \begin{array}{c} \mathrm{NP} \\ \mathrm{co\text{-}NP} \end{array} \subseteq \mathrm{PP} \subseteq \begin{array}{c} \mathrm{NP^{PP}} \\ \mathrm{co\text{-}NP^{PP}} \end{array} \subseteq \mathrm{PSPACE} \subseteq \mathrm{EXP}.$$

Because P is properly contained in EXP, EXP-complete problems are provably intractable; the other classes may equal P, although that is not generally believed to be the case.

Several other observations are worth making here. It is also known that $\mathrm{PH} \subseteq \mathrm{NP^{PP}}$, where PH represents the polynomial hierarchy. In a crude sense, PH is close to PSPACE, and, thus, our $\mathrm{NP^{PP}}$–completeness results place important problems close to PSPACE. However, some early empirical results (Littman, 1997b) show that random problem instances from PP have similar properties to random problem instances from NP, suggesting that PP might be close enough to NP for NP-type heuristics to be effective.

## 1.5 Results Summary

Tables 1 and 2 summarize our results, which are explained in more detail in later sections.

The general flavor of our main results and techniques can be conveyed as follows. To show that a plan-evaluation problem is in a particular complexity class $\mathcal{C}$, we take the cross product of the steps of the plan and the states of the domain and then look at the complexity of evaluating the absorption probability of the resulting Markov chain (i.e., the directed graph with probability-labeled edges). The complexity of the corresponding plan-existence problem is then bounded by $\mathrm{NP}^{\mathcal{C}}$, because the problem can be solved by guessing the correct plan non-deterministically and then evaluating it; in many cases, it is $\mathrm{NP}^{\mathcal{C}}$-complete. The appropriate complexity class $\mathcal{C}$ depends primarily on the representation of the cross-product Markov chain.

Exceptions to this basic pattern are the results for partially ordered plans in Section 4. These appear to require a distinct set of techniques.

It is also worth noting that, although propositional domains can be exponentially more compact than flat domains, the computational complexity of solving problems in propositional domains is not always exponentially greater; in one instance, evaluating partially ordered plans under the average interpretation, the complexity is actually the same for flat and propositional domains!

We also prove results concerning plan evaluation and existence for compactly represented plans (PP-complete and $\mathrm{NP^{PP}}$-complete, Corollary 5), plan existence of "large enough" looping plans in flat domains (P-complete, Theorem 7), plan evaluation and existence for looping plans in deterministic propositional domains (PSPACE-complete, Theorems 8 and 9), and plan existence for polynomial-size looping plans in partially observable domains (NP-complete, Section 5.1).

| Plan Type | Plan Evaluation | Plan Existence | Reference |
|---|---|---|---|
| unrestricted | — | P-complete | P & T (1987) |
| polynomial-depth | — | P-complete | P & T (1987) |
| looping | PL-complete | NP-complete | Section 3 |
| acyclic | PL-complete | NP-complete | Section 2 |
| totally ordered | PL-complete | NP-complete | Section 2 |
| partially ordered, optimistic | NP-complete | NP-complete | Section 4 |
| partially ordered, average | PP-complete | NP-complete | Section 4 |
| partially ordered, pessimistic | co-NP-complete | NP-complete | Section 4 |

Table 1: Complexity results for flat representations (P & T (1987) is Papadimitriou and Tsitsiklis (1987))

| Plan Type | Plan Evaluation | Plan Existence | Reference |
|---|---|---|---|
| unrestricted | — | EXP-complete | Littman (1997a) |
| polynomial-depth | — | PSPACE-complete | Littman (1997a) |
| looping | PSPACE-complete | PSPACE-complete | Section 3 |
| acyclic | PP-complete | $NP^{PP}$-complete | Section 2 |
| totally ordered | PP-complete | $NP^{PP}$-complete | Section 2 |
| partially ordered, optimistic | $NP^{PP}$-complete | $NP^{PP}$-complete | Section 4 |
| partially ordered, average | PP-complete | $NP^{PP}$-complete | Section 4 |
| partially ordered, pessimistic | co-$NP^{PP}$-complete | $NP^{PP}$-complete | Section 4 |

Table 2: Complexity results for propositional representations

## 2. Acyclic Plans

In this section, we treat the complexity of generating and evaluating acyclic and totally ordered plans.

**Theorem 1** *The plan-evaluation problem for acyclic and totally ordered plans in flat domains is* PL-*complete.*

**Proof:**  First, we show PL-hardness for totally ordered plans. Jung (1985) proved that a set $A$ is in PL if and only if there exists a logarithmically space-bounded and polynomially time-bounded nondeterministic Turing machine $N$ with the following property: For every input $x$, machine $N$ must have at least half of its computations on input $x$ be accepting if and only if $x$ is in $A$. The machine $N$ can be transformed into a probabilistic Turing machine $R$ such that for each input $x$, the probability that $R(x)$ accepts $x$ equals the fraction of computations of $N(x)$ that accepted. Given $R$, a planning domain $M$ can be described as follows. The state set of $M$ is the set of configurations of $R$ on input $x$. Note that a configuration consists of the contents of the logarithmically space-bounded tape, the state, the location of the read/write heads, and one symbol each from the input and output tapes. Thus, a configuration can be represented with logarithmically many bits, and there are only polynomially many such configurations. The state-transition probabilities of $M$ under the unique action $a$ are the configuration transition probabilities of $R$. All states obtained from accepting configurations are goal states. The totally ordered plan consists of a "step counter" for $R$ on input $x$, and each of its plan steps takes the only action $a$. The probability that the planning domain under this plan reaches a goal state is exactly the probability that $R(x)$ reaches an accepting configuration. Thus, evaluating this totally ordered plan is PL-hard.

Since totally ordered plans are acyclic plans, this also proves PL-hardness of the plan-evaluation problem for acyclic plans.

Next, we show that the plan-evaluation problem is in PL for acyclic plans. Let $M = \langle \mathcal{S}, s_0, \mathcal{A}, t, \mathcal{G} \rangle$ be a planning domain, let $P = \langle V, v_0, E, \pi, \delta \rangle$ be an acyclic plan, and let threshold $\theta$ be given. We show how our question, whether the probability that $M$ under $P$ reaches a goal state with probability greater than $\theta$, can be equivalently transformed into the question of whether a GapL function is greater than 0. The transformation can be done in logarithmic space. As shown by Allender and Ogihara (1996), it follows that our question is in PL.

At first, we construct a Markov chain $C$ from $M$ and $P$, which simulates the execution or "evaluation" of $M$ under $P$. Note that a Markov chain can be seen as a probabilistic domain with only one action in its set of actions. Since there is no choice of actions, we do not mention them in this construction. The state space of $C$ is $\mathcal{S} \times V$, the initial state is $(s_0, v_0)$, the set of goal states is $\mathcal{G} \times V$, and the transition probabilities $t_C$ for $C$ are

$$t_C((s,v),(s',v')) = \begin{cases} t(s, \pi(v), s'), & \text{if } s' \in \delta(v, v'), \\ 1, & \text{if } v \text{ is a terminal step node, and } (s,v) = (s',v'), \\ 0, & \text{otherwise.} \end{cases}$$

Let $m$ be the number of plan steps of $P$ (i.e., $|V|$, the number of nodes in the graph representing $P$). Since states of $C$ that contain a terminal step of $P$ are sinks in $C$, it follows

that

Pr($M$ reaches a goal state under $P$) = Pr($C$ reaches a goal state in exactly $m$ steps).

Let

$p_C(s, m) :=$ Pr($C$ reaches a goal state in exactly $m$ steps from initial state $s$).

Then, $p_C((s_0, v_0), m)$ is the probability we want to calculate. The standard inductive definition of $p_C$ used to evaluate plans by dynamic programming is

$$
\begin{aligned}
p_C(s, 0) &= \begin{cases} 1, & \text{if } s \text{ is a goal state of } C, \\ 0, & \text{otherwise,} \end{cases} \\
p_C(s, k+1) &= \sum_{s' \in \mathcal{S} \times V} t_C(s, s') \cdot p_C(s', k), \quad 0 \le k \le m-1.
\end{aligned}
$$

Let $h$ be the maximum length of the representation of a state-transition probability $t_C$. Then, for

$$
\begin{aligned}
p_h(s, 0) &= \begin{cases} 1, & \text{if } s \text{ is a goal state of } C, \\ 0, & \text{otherwise,} \end{cases} \\
p_h(s, k+1) &= \sum_{s' \in \mathcal{S} \times V} 2^h \cdot t_C(s, s') \cdot p_h(s', k), \quad 0 \le k \le m-1,
\end{aligned}
$$

it follows that $p_C((s_0, v_0), m) = p_h((s_0, v_0), m) \cdot 2^{-hm}$. Note that $p_h((s_0, v_0), m)$ is an integer value. Therefore, $p_C((s_0, v_0), m) > \theta$ if and only if $p_h((s_0, v_0), m) - \lfloor 2^{hm}\theta \rfloor > 0$. In order to show that $p_C((s_0, v_0), m) > \theta$ is decidable in PL, it suffices to show that $p_h((s_0, v_0), m)$ is in GapL. Therefore, we "unwind" the inductive definition of $p_h$. Let $T$ be the integer matrix obtained from $t_C$ with $T_{(s,s')} = t_C(s, s') \cdot 2^h$. We introduce the integer-valued $T$ to show that $p_h$ can be composed from GapL functions using compositions under which GapL is closed; as $t_C$ is not integer valued, it cannot be used to show this. We can write

$$
p_h(s, m) = \sum_{s' \in \mathcal{S} \times V} (T^m)_{(s,s')} \cdot p_h(s', 0).
$$

We argue that $p_h$ is in GapL. Each entry $T_{(s,s')}$ is logspace computable from the domain $M$ and plan $P$. Therefore, the powers of the matrix are in GapL, as shown by Vinay (1991). Because GapL is closed under multiplication and summation of polynomially many summands, it follows that $p_h \in$ GapL. Finally, we use closure properties of GapL from Allender and Ogihara (1996); since GapL is closed under subtraction, it follows that the plan-evaluation for acyclic plans is in PL.

Because totally ordered plans are acyclic plans, the plan-evaluation problem for totally ordered plans is also in PL. ∎

The technique of forming a Markov chain by taking the cross product of a domain and a plan will be useful later. Plan-existence problems require a different set of techniques.

**Theorem 2** *The plan-existence problem for acyclic and totally ordered plans in flat domains is* NP-*complete.*

**Proof:** First, we show containment in NP. Given a planning domain $M$, a threshold $\theta$, and a size bound $z \leq |M|$, guess a plan of the correct form of size at most $z$ and accept if and only if $M$ reaches a goal state with probability greater than $\theta$ under this plan. Note that checking whether a plan has the correct form can be done in polynomial time. Because the plan-evaluation problem is in PL (Theorem 1), it follows that the plan-existence problem is in NP (i.e., it is in $\text{NP}^{\text{PL}} = \text{NP}$).

To show the NP-hardness of the plan-existence problem, we give a reduction from the NP-complete satisfiability problem for Boolean formulae in conjunctive normal form. We construct a planning domain that evaluates a Boolean formula with $n$ variables, where a $(n+2)$-step plan describes an assignment of values to the variables. In the first step, a clause is chosen randomly. At step $i+1$, the planning domain "checks" whether the plan satisfies the appearance of variable $i$ in that clause. If so, the clause is marked as satisfied. After $n+1$ steps, if no literal was satisfied in that clause, then no goal state is reached through this clause, otherwise, a transition is made to the goal state. Therefore, the goal state will be reached with probability 1 (greater than $1 - 1/m$) if and only if all clauses are satisfied—the plan describes a satisfying assignment.

We formally define the reduction, which is similar to one presented by Papadimitriou and Tsitsiklis (1987). Let $\phi$ be a CNF formula with $n$ variables $x_1, \ldots, x_n$ and $m$ clauses $C_1, \ldots, C_m$. Let the sign of an appearance of a variable in a clause be $-1$ if the variable is negated, and 1 otherwise. Define the planning domain $M(\phi) = \langle \mathcal{S}, s_0, \mathcal{A}, t, \mathcal{G} \rangle$ where

$$
\begin{aligned}
\mathcal{S} &= \{\text{sat}(i,j), \text{unsat}(i,j) \mid 1 \leq i \leq n+1, 1 \leq j \leq m\} \cup \{s_0, s_{\text{acc}}, s_{\text{rej}}\}, \\
\mathcal{A} &= \{\text{assign}(i,b) \mid 1 \leq i \leq n, b \in \{-1, 1\}\} \cup \{\text{start}, \text{end}\}, \\
\mathcal{G} &= \{s_{\text{acc}}\},
\end{aligned}
$$

$$
t(s,a,s') = \begin{cases}
\frac{1}{m}, & \text{if } s = s_0, a = \text{start}, s' = \text{unsat}(1,j), 1 \leq j \leq m, \\
1, & \text{if } s = s_0, a \neq \text{start}, s' = s_{\text{rej}}, \\
1, & \text{if } s = \text{unsat}(i,j), a = \text{assign}(i,b), s' = \text{sat}(i+1,j), i \leq n, \\
& \quad x_i \text{ appears in } C_j \text{ with sign } b, \\
1, & \text{if } s = \text{unsat}(i,j), a = \text{assign}(i,b), s' = \text{unsat}(i+1,j), i \leq n, \\
& \quad x_i \text{ does not appear in } C_j \text{ with sign } b, \\
1, & \text{if } s = \text{unsat}(i,j), a = \text{assign}(i',b) \text{ or } a = \text{start or } a = \text{end}, \\
& \quad s' = s_{\text{rej}}, i' \neq i \leq n, b \in \{-1, 1\}, \\
1, & \text{if } s = \text{unsat}(n+1,j), s' = s_{\text{rej}}, \\
1, & \text{if } s = \text{sat}(i,j), a = \text{assign}(i,b), s' = \text{sat}(i+1,j), i \leq n, \\
1, & \text{if } s = \text{sat}(i,j), a = \text{assign}(i',b) \text{ or } a = \text{start or } a = \text{end}, \\
& \quad s' = s_{\text{rej}}, i' \neq i \leq n, \\
1, & \text{if } s = \text{sat}(n+1,j), a = \text{end}, s' = s_{\text{acc}}, \\
1, & \text{if } s = \text{sat}(n+1,j), a \neq \text{end}, s' = s_{\text{rej}}, \\
1, & \text{if } s = s' = s_{\text{rej}} \text{ or } s = s' = s_{\text{acc}}, \\
0, & \text{otherwise.}
\end{cases}
$$

The meaning of the states in this domain is as follows. When the domain is in state $\text{sat}(i,j)$ for $1 \leq i \leq n$, $1 \leq j \leq m$, it means the formula has been satisfied, and we are currently checking variable $i$ in clause $j$. State $\text{sat}(n+1,j)$ for all $1 \leq j \leq m$ means that we've finished verifying clause $j$ and it was satisfied. The meanings are similar for the "unsat" states. Of course, $s_0$ is the initial state and $s_{\text{acc}}$ and $s_{\text{rej}}$ are the accepting and rejecting states, respectively.

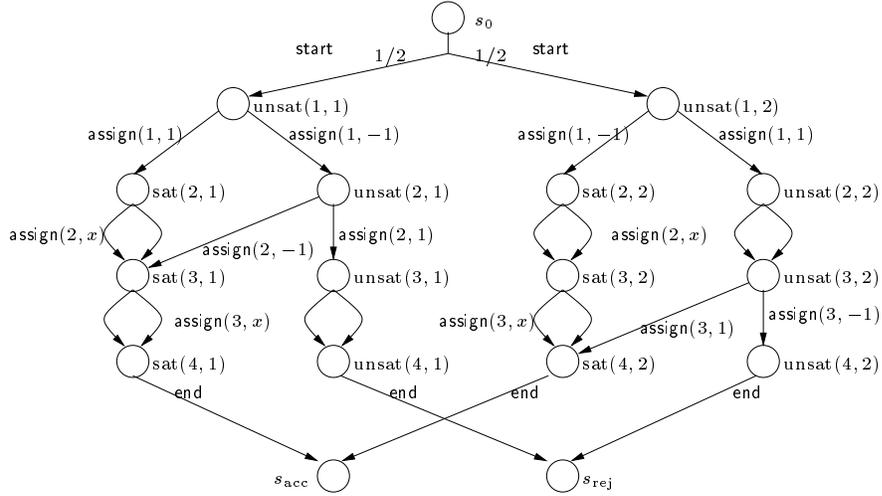

Figure 3: A domain generated from the Boolean formula $(x_1 \vee \neg x_2) \wedge (\neg x_1 \vee x_3)$

The actions in this domain are start and end, which mark the beginning and end of the assignment, and assign($i, b$) for $1 \le i \le n$, $b \in \{-1, 1\}$, which assign the truth value $b$ to variable $i$. Figure 3 gives the domain generated by this reduction from a simple Boolean formula. By the description of the reduction, $M(\phi)$ can be computed from $\phi$ in time polynomial in $|\phi|$.

By construction, $M(\phi)$ under $z = (n + 2)$-step plan $P$ can only reach goal state $s_{\text{acc}}$ if $P$ has the form

$$\text{start} \rightarrow \text{assign}(1, b_1) \rightarrow \text{assign}(2, b_2) \rightarrow \cdots \rightarrow \text{assign}(n, b_n) \rightarrow \text{end} \rightarrow \bullet.$$

$P$ reaches $s_{\text{acc}}$ with probability 1 if and only if $b_1, \ldots, b_n$ is a satisfying assignment for the $n$ variables in $\phi$. This shows that Boolean satisfiability polynomial-time reduces to the plan-existence problem for totally ordered and acyclic plans, showing that it is NP-hard. ∎

Note that if we bound the plan depth (horizon) instead of the plan size, the plan-existence problem for acyclic plans in flat domains is P-complete (Goldsmith et al., 1997a; Papadimitriou & Tsitsiklis, 1987). Limiting the plan size makes the problem more difficult because it is possible to force the planner to take the same action from different states; figuring out how to do this without sacrificing plan quality is very challenging.

In propositional domains, plan evaluation is harder because of the large number of states.

**Theorem 3** *The plan-evaluation problem for acyclic and totally ordered plans in propositional domains is* PP-complete.

**Proof:** To show PP-hardness for totally ordered plans, we give a reduction from the PP-complete problem MAJSAT: given a CNF Boolean formula $\phi$, does the majority of assignments satisfy it?

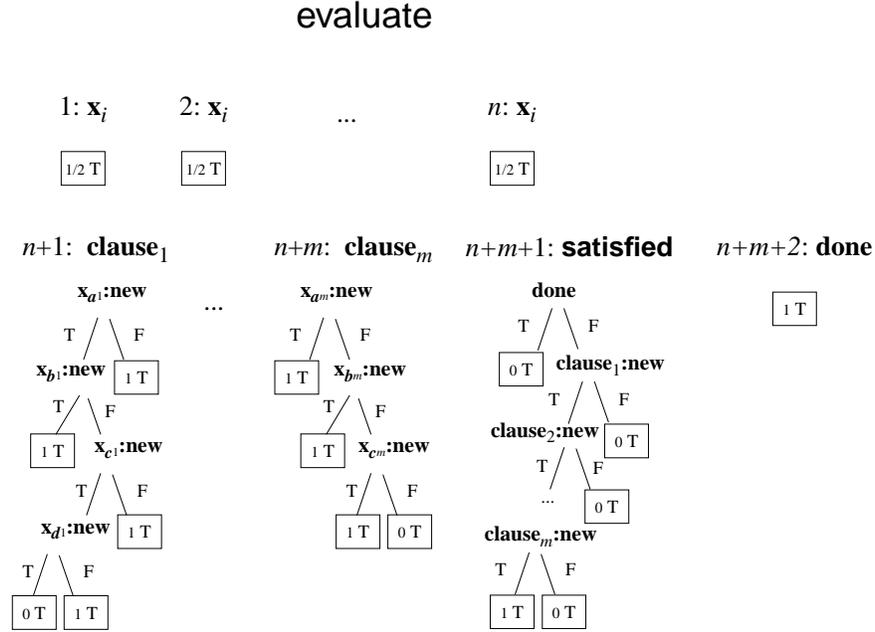

Figure 4: Sequential-effects-tree representation for evaluate

Given $\phi$, we construct a planning domain $\mathbb{M}(\phi)$ and a 1-step plan such that the plan achieves the goal with probability greater than $\theta = 1/2$ if and only if the majority of assignments satisfies $\phi$. The planning domain $\mathbb{M}(\phi)$ consists of a single action evaluate, which is also the 1-step plan to be evaluated. There are $n + m + 2$ propositions in $\mathbb{M}(\phi)$; $\mathbf{x}_1$ through $\mathbf{x}_n$, which correspond to the $n$ variables of $\phi$; $\mathbf{clause}_1$ through $\mathbf{clause}_m$, which correspond to the $m$ clauses of $\phi$; $\mathbf{satisfied}$, which is also the sole element of the goal set; and $\mathbf{done}$, which insures that evaluate is only executed once (this is important when this domain is used later in Theorem 4 to show the complexity of plan existence). In the initial state, all propositions are false.

The evaluate action generates a random assignment to the variables of $\phi$, evaluates the clauses ($\mathbf{clause}_i$ is true if any of the literals in the $i$th clause is true), and evaluates the entire formula ($\mathbf{satisfied}$ is true if all the clauses are true). Figure 4 gives an ST representation of evaluate, in which $x_{a^i}, x_{b^i}, \ldots$ represent the variables in clause $i$.

By construction, $\phi$ is in MAJSAT if and only if $\mathbb{M}(\phi)$ reaches a goal state with probability greater than $\theta = 1/2$ under the plan consisting of the single action evaluate.

We next show membership in PP for acyclic plans. We do this by showing that a planning domain $\mathbb{M}$ and an acyclic plan $P$ induce a computation tree consisting of all paths through $\mathbb{M}$ under $P$. Evaluating this computation tree can be accomplished by a PP machine.

Let $b$ be a bound on the number of bits used to specify probabilities in the leaves of the decision trees representing $\mathbb{M}$.[4] Consider a *computation tree* defined as follows. It has root labeled $\langle s_0, v_0 \rangle$. If, in the planning domain $\mathbb{M}$, the probability of reaching state $s'$ from $s$ given action $\pi(v)$ is equal to $\nu$, then $\langle s, \pi(v) \rangle$ will have $\nu \cdot 2^b$ children labeled $\langle s', \delta(v, s') \rangle$. Each of the identically labeled child nodes is independent but is defined identically to the others. Thus, the number of paths with a given set of labels corresponds to the probability of that trajectory through the domain and plan multiplied by $(2^b)^h$, where $h$ is the depth of the plan.

---

4. We represent numbers in polynomial-precision binary representation. In principle, this could introduce round-off errors if planning problems are specified in some other form.

The number of accepting computations is, therefore, more than $\theta \cdot (2^b)^h$ if and only if the probability of achieving the goal is more than $\theta$. Note that $b$ is inherent in the planning domain, rather than in $h$. A PP machine accepts if more than half of the final states are accepting, so if $\theta \neq 1/2$, it will be necessary to pad the computation tree by introducing "dummy" branches that accept or reject in the right proportions. ∎

The plan-existence problem is essentially equivalent to guessing and evaluating a valid plan.

**Theorem 4** *The plan-existence problem for acyclic and totally ordered plans in propositional domains is $\mathrm{NP}^{\mathrm{PP}}$-complete.*

**Proof:** Containment in $\mathrm{NP}^{\mathrm{PP}}$ for both totally ordered and for acyclic plans follows from the fact that a polynomial-size plan can be guessed in polynomial time and checked in PP (Theorem 3).

Hardness for $\mathrm{NP}^{\mathrm{PP}}$ for both totally ordered and acyclic plans can be shown using a reduction from E-MAJSAT, shown $\mathrm{NP}^{\mathrm{PP}}$-hard in the Appendix. The reduction echoes the one used in the PP-hardness argument in the proof of Theorem 3.

Given a CNF Boolean formula $\phi$ with variables $x_1, \ldots, x_n$, and a number $k$, we construct a planning domain $\mathbb{M}(\phi, k)$ such that a plan exists that can reach the goal with probability greater than $\theta = 1/2$ if and only if there is an assignment to the variables $x_1, \ldots, x_k$ such that the majority of assignments to the remaining variables satisfies $\phi$. The planning domain $\mathbb{M}(\phi, k)$ consists of the action evaluate from Theorem 3 and one action, set-$x_i$, for each of the first $k$ variables. Just as in the proof of Theorem 3, there are $n + m + 2$ propositions in $\mathbb{M}(\phi, k)$, all initially false: $\mathbf{x}_1$ through $\mathbf{x}_n$, which correspond to the $n$ variables of $\phi$; $\mathbf{clause}_1$ through $\mathbf{clause}_m$, which correspond to the $m$ clauses of $\phi$; $\mathbf{satisfied}$; and $\mathbf{done}$, which insures that evaluate is only executed once. The goal set contains $\mathbf{satisfied}$ and $\mathbf{done}$.

For $1 \leq i \leq k$, action set-$x_i$ makes proposition $\mathbf{x}_i$ true. Analogously to Theorem 3, the evaluate action generates a random assignment to the remaining variables of $\phi$, evaluates the clauses ($\mathbf{clause}_i$ is true if any of the literals in the clause is true), and evaluates the entire formula ($\mathbf{satisfied}$ is true if all the clauses are true), and sets $\mathbf{done}$ to true. If $\mathbf{done}$ is true, no further action can make $\mathbf{satisfied}$ true.

If the pair $\phi, k$ is in E-MAJSAT, then there exists an assignment $b_1 \ldots b_k$ to the first $k$ variables of $\phi$ such that the majority of assignments to the rest of the variables satisfies $\phi$. Therefore, the plan applying steps set-$x_i$ for all $i$ with $b_i = 1$ followed by an evaluate action reaches a goal state with probability greater than $\theta = 1/2$.

Conversely, assume $\mathbb{M}(\phi, k)$ under totally ordered plan $P$ reaches a goal state with probability greater than $1/2$. Since the evaluate action is the only action setting $\mathbf{done}$ to true, and since no action reaches the goal once $\mathbf{done}$ is set to true, we can assume without loss of generality that $P$ consists of a sequence of steps set-$x_i$ that ends with evaluate. By construction, the assignment to $x_1, \ldots, x_k$ assigning 1 exactly to those variables set by $P$ is an assignment under which the majority of the assignments to the rest of the variables satisfies $\phi$, and therefore $\phi, k$ is in E-MAJSAT.

Since every totally ordered plan is acyclic, the same hardness holds for acyclic plans. ∎

In the above results, we consider both flat and compactly represented (propositional) planning domains but only flat plans. Compactly represented plans are also quite useful.

> A *compact acyclic plan* is an acyclic plan in which the names of the plan steps are encoded by a set of propositional variables and the step-transition function $\delta$ between plan steps is represented by a set of decision trees, just as in ST. We require that the plan has depth polynomial in the size of the representation, even though the total number of steps in the plan might be exponential due to the logarithmic succinctness of the encodings.

Because the plan-domain cross-product technique used in the proof of Theorem 3 generalizes to compact acyclic plans, the same complexity results apply. This also holds true for a *probabilistic acyclic plan*, which is an acyclic plan that can make random transitions between plan steps (i.e., the step-transition function $\delta$ is stochastic). These insights can be combined to yield the following corollary of Theorems 3 and 4.

**Corollary 5** *The plan-evaluation problem for compact probabilistic acyclic plans in propositional domains is* PP-*complete and the plan-existence problem for compact probabilistic acyclic plans in propositional domains is* NP$^{\text{PP}}$-*complete.*

We mention probabilistic plans here for two reasons. First, the behavior of some planning structures (such as partially ordered plan evaluation under the average interpretation, discussed in Section 4) can be thought of as generating probabilistic plans. Second, there are many instances in which simple probabilistic plans perform nearly as well as much larger and more complicated deterministic plans; this notion is often exploited in the field of randomized algorithms. Work by Platzman (1981) (described by Lovejoy, 1991) shows how the idea of randomized plans can come in handy for planning in partially observable domains.

## 3. Looping Plans

Looping plans can be applied to infinite-horizon control. The complexity of plan existence and plan evaluation in flat domains (Theorems 1 and 2) does not depend on the presence or absence of loops in the plan.

**Theorem 6** *The plan-evaluation problem for looping plans in flat domains is* PL-*complete.*

**Proof:** Given a domain $M$ and a looping plan $P$, we can construct a product Markov chain $C$ as in the proof of Theorem 1. As in the proof of Theorem 6 of Allender and Ogihara (1996), this chain can be constructed such that it has exactly one accepting and exactly one rejecting state; both of these states are absorbing. The probability that $M$ reaches a goal state under $P$ equals the probability that $C$ reaches its accepting state if started in its initial state, which is the product of the initial states of $M$ and $P$. In the proof of Theorem 6 of Allender and Ogihara (1996), it is shown that the construction of the Markov chain and the computation of whether it reaches its final state with probability greater than $\theta$ can be performed in PL.

PL-hardness is implied by Theorem 1, since acyclic plans are a special case of looping plans. ∎

**Theorem 7** *The plan-existence problem for looping plans in flat domains is* NP-*complete in general, but* P-*complete if the size of the desired plan is at least the size of the state or action space (i.e., $z \geq \min(|\mathcal{S}|, |\mathcal{A}|)$).*

**Proof sketch:** NP-completeness follows from the proof of Theorem 2; containment and hardness still hold if plans are permitted to be looping.

However, this is only true if we are forced to specify a plan whose size is small with respect to the size of the domain. If our looping plan is allowed to have a number of states that is at least as large as the number of states or actions in the domain, the problem can be solved in polynomial time.

It is known that for Markov decision processes such as these the maximum probability of reaching a goal state equals the maximum probability of reaching a goal state under any infinite-horizon *stationary policy*, where a stationary policy is a mapping from states to actions that is used repeatedly to choose actions at each time step. It is known that such an optimal stationary policy can be computed in polynomial time via linear programming (Condon, 1992). Any stationary policy for a domain $M = \langle \mathcal{S}, s_0, \mathcal{A}, \mathcal{G}, t \rangle$ can be written as a looping plan, although, of course, not all looping plans correspond to stationary policies.

We show that for any fixed stationary policy $p : \mathcal{S} \to \mathcal{A}$, there are two simple ways a looping plan $P = (V, v_0, E, \pi, \delta)$ can be represented. First, let $V = \mathcal{A}$, $v_0 = p(s_0)$, $\pi(v) = v$, and $\delta(v, v') = \{s \in \mathcal{S} \mid p(s) = v'\}$. It follows that whenever $M$ reaches state $s$, then the action applied according to the looping plan is the same as according to $P$.

Second, let $V = \mathcal{S}$, $v_0 = s_0$, $\pi(v) = p(v)$, and $\delta(v, v') = \{v'\}$. It follows that whenever $M$ reaches state $s$, the plan will be at the node corresponding to that state and, therefore, the appropriate action for that state will be applied by the looping plan. Therefore, the maximum probability of reaching a goal state can be obtained by either of these looping plans.

Since the best stationary policy can be computed in polynomial time, the best looping plan can be computed in polynomial time, too. P-hardness follows from a theorem of Papadimitriou and Tsitsiklis (1987). ∎

In propositional domains, the complexity of plan existence and plan evaluation of looping plans is quite different from the acyclic case. Looping plan evaluation is very hard.

**Theorem 8** *The plan-evaluation problem for looping plans in both deterministic and stochastic propositional domains is* PSPACE-*complete.*

**Proof:** Recall that the plan-evaluation problem for flat domains is in PL (Theorem 1). For a planning domain with $c^n$ states and a representation of size $n$, a looping plan can be evaluated in probabilistic space $\mathcal{O}(\log(c^n))$ (Theorem 6), which is to say probabilistic space polynomial in the size of the input. This follows because the ST representation of the domain can be used to compute entries of the transition function $t$ in polynomial space. Since probabilistic PSPACE equals PSPACE, this shows that the plan-evaluation problem for looping plans in stochastic propositional domains is in PSPACE.

It remains to show PSPACE-hardness for deterministic propositional domains. Let $N$ be a deterministic polynomial-space-bounded Turing machine. The moment-to-moment computation state (configuration) of $N$ can be expressed as a polynomial-length bit string that encodes the contents of the Turing machine's tape, the location of the read/write head, the state of $N$'s finite-state controller, and whether or not the machine is in an accepting state.

For any input $x$, we describe how to construct in polynomial time a deterministic planning domain $\mathbb{M}(x)$ and a single-action looping plan that reaches a goal state of $\mathbb{M}(x)$ if and only $x$ is accepted by Turing machine $N$.

Given a description of $N$ and $x$, one can, in time polynomial in the size of the descriptions of $N$ and $x$, produce a description of a Turing machine $T$ that computes the transition function for $N$. In other words, $T$ on input $c$, a configuration of $N$, outputs the next configuration of $N$. (In fact, $T$ can even check whether $c$ is a valid configuration in the computation of $N(x)$ by simulating that computation.) By an argument similar to that used in Cook's theorem, $T$ can be modeled by a polynomial-size circuit. This circuit takes as input the bit string describing the current configuration of $N$ and outputs the next configuration.

Next, we argue that the computation of this circuit can be expressed by an action compute in ST representation. There is one proposition in $\mathbb{M}(x)$ for each bit in the configuration, plus one for each gate of the circuit. The three standard gates, "and," "or," and "not" are all easily represented as decision trees. By ordering the decision trees in compute according to a topological sort of the gates of the circuit, a single compute action can compute precisely the same output as the circuit. Figure 5 illustrates this conversion for a simple circuit, which gives the form of the "not" $(i_1)$, "and" $(i_2)$, and "or" $(i_3)$ gates.

We can now describe the complete reduction. The planning domain $\mathbb{M}(x)$ consists of the single action compute and the set of propositions described in the previous paragraph. The initial state is the initial configuration of the Turing machine $N$, and the goal set is the proposition corresponding to whether or not the configuration is an accepting state for $N$.

Because all transitions are deterministic and only one action can be chosen, it follows that the goal state is reached with probability 1 (greater than $1/2$, for example) under the plan that repeatedly chooses compute until an accepting state is reached if and only if polynomial-space machine $N$ on input $x$ accepts. ∎

A similar argument shows that looping plan existence is not actually any harder than looping plan evaluation.

**Theorem 9** *The plan-existence problem for looping plans in both deterministic and stochastic propositional domains is* PSPACE-complete.

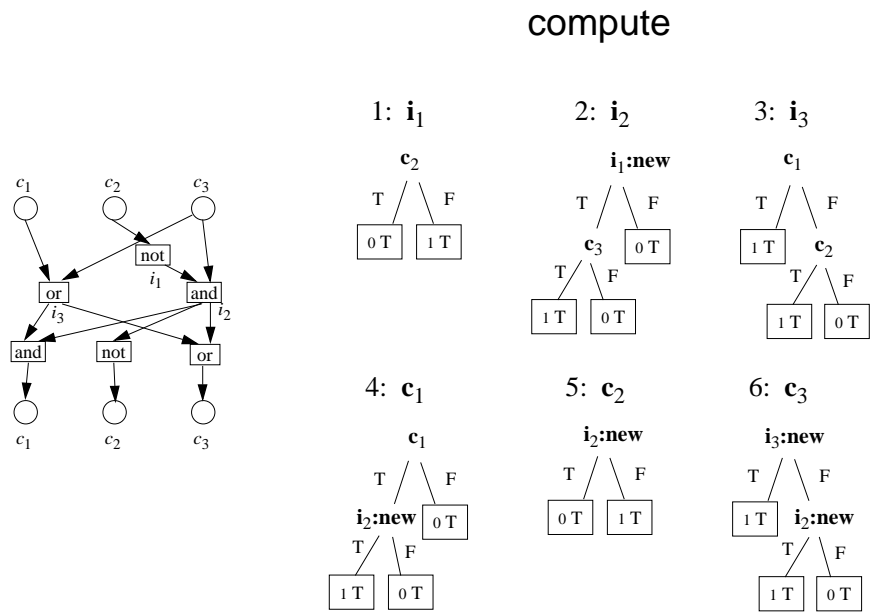

Figure 5: A circuit and its representation as a sequential-effects tree

**Proof:** Hardness for PSPACE follows from the same construction as in the proof of Theorem 8: either the one-step looping plan is successful, or it is not. No other plan yields a better result.

Recall that we are only interested in determining whether there is a plan of size $z$, where $z$ is bounded by the size of the domain, that reaches the goal with a given probability. The problem is in PSPACE because the plan can be guessed in polynomial time and checked in PSPACE (Theorem 8). Because $\text{NP}^{\text{PSPACE}} = \text{PSPACE}$, the result follows. ∎

As we mentioned earlier, the *unrestricted* infinite-horizon plan-existence problem is EXP-complete (Littman, 1997a); this shows the problem of determining unrestricted plan existence is EXP-hard only because some domains require plans that are larger than polynomial-size looping plans.

Because Theorem 9 shows PSPACE-completeness for determining plan existence in deterministic domains, it is closely related to the PSPACE-completeness result of Bylander (1994). The main difference between the two results is that our theorem applies to more compact plans (polynomial instead of exponential) with more complex operator descriptions (conditional effects instead of preconditions with add and delete lists) that can include loops. Also, as the proofs above show, PSPACE-hardness is retained even in planning domains with only one action, so it is the looping that makes looping plans hard to work with.

## 4. Partially Ordered Plans

Partially ordered plans are a popular representation because they allow planning algorithms to defer a precise commitment to the ordering of plan steps until it becomes necessary in the planning process. A $k$-step partially ordered plan corresponds to a set of $k$-step totally ordered plans—all those that are consistent with the given partial order. The evaluation of a partially ordered plan can be defined to be the evaluation of the best, worst, or average member of the set of consistent totally ordered plans; these are the optimistic, pessimistic, and average interpretations, respectively.

The plan-evaluation problem for partially ordered plans is different from that of totally ordered plans. This is because a single partial order can encode all totally ordered plans. Hence, evaluating a partially ordered plan involves figuring out the best (in case of optimistic interpretation) or the worst (for pessimistic interpretation) member, or the average (for average interpretation) of this combinatorial set.

**Theorem 10** *The plan-evaluation problem for partially ordered plans in flat domains is* NP-*complete under the optimistic interpretation.*

**Proof sketch:** Membership in NP follows from the fact that we can guess any totally ordered plan consistent with the given partial order and accept if and only if the domain reaches a goal state with probability more than $\theta$. Remember that this evaluation can be performed in PL (Theorem 1), and therefore deterministically in polynomial time.

The hardness proof is a variation of the construction used in Theorem 2. The partially-ordered plan to evaluate has the form given in Figure 6; the consistent total orders are of the form

$$\mathsf{start} \to \mathsf{assign}(1, b_1) \to \mathsf{assign}(1, -b_1) \to \mathsf{assign}(2, b_2) \to \mathsf{assign}(2, -b_2) \to$$

$$\cdots \to \mathsf{assign}(n, b_n) \to \mathsf{assign}(n, -b_n) \to \mathsf{end} \to \bullet,$$

where $b_i$ is either 1 or $-1$. Each of the possible plans can be interpreted as an assignment to $n$ Boolean variables by ignoring every second assignment action. The construction in Theorem 2 shows how to turn a CNF formula $\phi$ into a planning domain $M(\phi)$, and it can easily be modified to ignore every second action. Thus, the best totally ordered plan consistent with the given partially ordered plan reaches the goal with probability 1 if and only if it reaches the goal with probability greater than $1 - 2^{-m}$ if and only if it satisfies all clauses of $\phi$ if and only if $\phi$ is satisfiable. ∎

**Theorem 11** *The plan-evaluation problem for partially ordered plans in flat domains is* co-NP-*complete under the pessimistic interpretation.*

**Proof sketch:** Both the proof of membership in co-NP and the proof of hardness are very similar to the proof of Theorem 10. We show a reduction from the co-NP-complete set $\overline{\mathrm{SAT}}$ of unsatisfiable formulae in CNF. The plan to evaluate has the form given in Figure 6 and is interpreted as above. As in the proof of Theorem 2, we construct a planning domain $M'(\phi)$, but we take $\mathcal{G} = \{s_{\mathrm{rej}}\}$ as goal states, where the state $s_{\mathrm{rej}}$ is reached with probability greater than 0 if and only if the assignment does not satisfy one of the clauses of formula $\phi$. A formula is unsatisfiable if and only if under every assignment at least one of the clauses is not satisfied. Therefore, the probability that $M'(\phi)$ reaches a goal state under a given totally ordered plan is greater than 0 if and only if the plan corresponds to an unsatisfying assignment. Finally, the minimum of that probability over all consistent partially ordered plans is greater than 0 if and only if $\phi$ is unsatisfiable. ∎

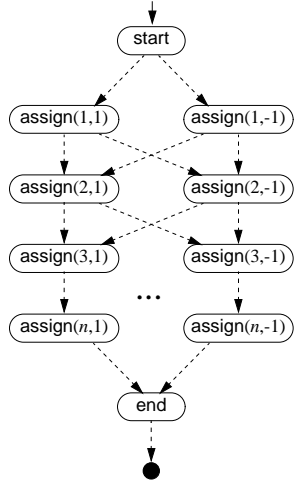

Figure 6: A partially ordered plan that can be hard to evaluate

**Theorem 12** *The plan-evaluation problem for partially ordered plans in flat domains is PP-complete under the average interpretation.*

**Proof:** Under the average interpretation, we must decide whether the average evaluation over all consistent totally ordered plans is greater than threshold $\theta$. This can be decided in PP by guessing uniformly a totally ordered plan and checking its consistency with the given partially ordered plan in polynomial time. If the guessed totally ordered plan is consistent, it can be evaluated in polynomial time (Theorem 1) and accepted or rejected as appropriate. If the guessed plan is inconsistent, the computation accepts with probability $\theta$ and rejects with probability $1 - \theta$, leaving the average over the consistent orderings unchanged with respect to the threshold $\theta$.

The PP-hardness is shown by a reduction from the PP-complete MAJSAT. Let $\phi$ be a formula in CNF. We show how to construct a domain $M(\phi)$ and a partially ordered plan $P(\phi)$ such that $\phi \in$ MAJSAT if and only if the average performance of $M(\phi)$ under a totally ordered plan consistent with $P(\phi)$ is greater than $1/2$.

Let $\phi$ consist of the $m$ clauses $C_1, \ldots, C_m$, which contain $n$ variables $x_1, \ldots, x_n$. Domain $M(\phi) = \langle \mathcal{S}, s_0, \mathcal{A}, t, \mathcal{G} \rangle$ has actions

$$\mathcal{A} = \{\mathsf{assign}(i, b) \mid i \in \{1, \ldots, n\}, b \in \{-1, 1\}\} \cup \{\mathsf{start}, \mathsf{check}, \mathsf{end}\}.$$

Action $\mathsf{assign}(i, b)$ will be interpreted as "assign sign $b$ to $x_i$." The partially ordered plan $P(\phi)$ has plan steps

$$V = \{\sigma(i, b, h) \mid i \in \{1, \ldots, n\}, b \in \{-1, 1\}, h \in \{1, \ldots, m\}\} \cup \{\mathsf{start}, \mathsf{check}, \mathsf{end}\}$$

and mapping $\pi : V \to \mathcal{A}$ with

$$\pi(\sigma) = \sigma \text{ for } \sigma \in \{\mathsf{start}, \mathsf{check}, \mathsf{end}\}, \text{ and } \pi(\sigma(i, b, h)) = \mathsf{assign}(i, b).$$

The order $E$ requires that a consistent plan has start as the first and end as the last step. The steps in between are arbitrarily ordered. More formally,

$$E = \{(\text{start}, q) \mid q \in V - \{\text{start}, \text{end}\}\} \cup \{(q, \text{end}) \mid q \in V - \{\text{start}, \text{end}\}\}.$$

Now, we define how the domain $M(\phi)$ acts on a given totally ordered plan $P$ consistent with $P(\phi)$. Domain $M(\phi)$ consists of the cross product of the following polynomial-size deterministic domains $M_s$ and $M_\phi$, to which a final probabilistic transition will be added.

Before we describe $M_s$ and $M_\phi$ precisely, here are their intuitive definitions. The domain $M_s$ is satisfied by plans that have the form of an assignment to the $n$ Boolean variables with the restriction that the assignment is repeated $m$ times (for easy checking). The domain $M_\phi$ is satisfied by plans that correspond to satisfying assignments. The composite of these two domains is only satisfied by plans that correspond to satisfying assignments. We will now define these domains formally.

First, $M_s$ checks whether the totally ordered plan matches the regular expression

$$\text{start } (\text{assign}(1, 0)^m | \text{assign}(1, 1)^m)$$
$$\cdots (\text{assign}(n, 0)^m | \text{assign}(n, 1)^m)$$
$$\text{check } ((\text{assign}(1, 0) | \text{assign}(1, 1)) \cdots (\text{assign}(n, 0) | \text{assign}(n, 1)))^m.$$

Note that the $m$ here is a constant. Let "good" be the state reached by $M_s$ if the plan matches that expression. Otherwise, the state reached is "bad". To clarify, the actions before check are there simply to "use up" the extra steps not used in specifying the assignment in the partially ordered plan.

Next, $M_\phi$ checks whether the sequence of actions following the check action satisfies the clauses of $\phi$ in the following sense. Let $a_1 \cdots a_k$ be this sequence. $M_\phi$ interprets each subsequence $a_{1+(j-1)n} \cdots a_{n+(j-1)n}$ with $a_{l+(j-1)m} = \text{assign}(x, b_l)$ as assignment $b_1, \ldots, b_n$ to the variables $x_1, \ldots, x_n$, and checks whether this assignment satisfies clause $C_j$. If all single clauses are satisfied in this way, then $M_\phi$ reaches state "satisfied".

Note that $M_s$ and $M_\phi$ are defined so that they do not deal with the final end action. $M(\phi)$ consists of the product domain of $M_s$ and $M_\phi$ with the transitions for action end as follows. If $M$ is in state (bad, $q$) for any state $q$ of $M_\phi$, then action end lets $M$ go probabilistically to state "accept" or to state "reject", with probability $1/2$ each; if $M$ is in state (good, satisfied), the $M$ under action end goes to state "accept" (with probability 1); otherwise, $M$ under action end goes to state reject (with probability 1). The set of goal states of $M$ consists of the only state "accept".

We analyze the behavior of $M(\phi)$ under any plan $P$ consistent with $P(\phi)$. If $M_s$ under $P$ reaches state "bad", then $M(\phi)$ under $P$ reaches a goal state with probability $1/2$. Now, consider a plan $P$ under which $M_s$ reaches the state "good"—called a *good* plan. Then $P$ matches the above regular expression. Therefore, for every $i \in \{1, \ldots, m\}$ there exists $b_i \in \{-1, 1\}$ such that all steps $s(i, b_i, h)$ are between start and check. Thus, all steps between check and end are

$$s(1, 1 - i_1, 1) \cdots s(n, 1 - i_n, 1) s(1, 1 - i_1, 2) \cdots s(n, 1 - i_n, m)$$

Consequently, the sequence of actions defined by the labeling of these plan steps are

$$(\text{assign}(1, i_1) \text{assign}(2, i_2) \cdots \text{assign}(n, i_n))^m.$$

This means, that $M_\phi$ checks whether all clauses of $\phi$ are satisfied by the assignment $i_1 \cdots i_n$, i.e., $M_\phi$ checks whether $i_1 \cdots i_n$ satisfies $\phi$. Therefore, $M(\phi)$ accepts under plan $P$ with probability 1, if the plan represents a satisfying assignment, and with probability 0 otherwise.

Note that each assignment corresponds to exactly one good plan. Therefore, the average over all good plans that $M(\phi)$ accepts equals the fraction of satisfying assignments of $\phi$. Since $M(\phi)$ accepts under "bad" plans with probability $1/2$, this yields that the average over all plans consistent with $P(\phi)$ of the acceptance probabilities of $M(\phi)$ is greater than $1/2$ if and only if $\phi \in$ Majsat. ∎

The complexity of the plan-existence problem for partially ordered plans is identical to that for totally ordered plans.

**Theorem 13** *The plan-existence problem for partially ordered plans in flat domains is* NP-*complete under the pessimistic, optimistic and average interpretations. The plan-existence problem for partially ordered plans in propositional domains is* $\mathrm{NP^{PP}}$-*complete under the pessimistic, optimistic and average interpretations.*

**Proof:** First, note that a totally ordered plan is a special type of partially ordered plan and its evaluation is unchanged under the pessimistic, optimistic, or average interpretation. In particular, because there is only one ordering consistent with a given totally ordered plan, the best, worst, and average orderings are all the same. Therefore, if there exists a totally ordered plan with value greater than $\theta$, then there is a partially ordered plan with value greater than $\theta$ (the same plan), under all three interpretations.

Conversely, if there is a partially ordered plan with value greater than $\theta$ under any of the three interpretations, then there is a totally ordered plan with value greater than $\theta$. This is because the value of the best, worst, and average ordering of a partially ordered plan is always a lower bound on the value of the best consistent totally ordered plan.

Given this strong equivalence, the complexity of plan existence for partially ordered plans is a direct corollary of Theorems 2 and 4. ∎

The pattern for partially ordered plan evaluation in flat domains is that the average interpretation is no easier to decide than either the optimistic or pessimistic interpretations. In propositional domains, the pattern is the opposite: the average interpretation is no harder to decide than either the optimistic or pessimistic interpretations.

**Theorem 14** *The plan-evaluation problem for partially ordered plans in propositional domains is* $\mathrm{NP^{PP}}$-*complete under the optimistic interpretation,* co-$\mathrm{NP^{PP}}$-*complete under the pessimistic interpretation, and* PP-*complete under the average interpretation.*

**Proof sketch:** For the optimistic interpretation, membership in $\mathrm{NP^{PP}}$ follows from the fact that we can guess a single sufficiently good consistent total order and evaluate it in PP (Theorem 3). Hardness for $\mathrm{NP^{PP}}$ can be shown using a straightforward reduction from E-Majsat (as in the proof of Theorem 4).

For the pessimistic interpretation, membership in co-$\mathrm{NP^{PP}}$ follows from the fact that we can guess the *worst* consistent total order and evaluate it in PP (Theorem 3). Hardness for

co-NP$^{PP}$ can be shown by reducing to it the co-NP$^{PP}$ version of E-Majsat ($\overline{\text{E-Majsat}}$); the proof is a simple adaptation of the techniques used, for example, in Theorem 4 above.

For the average interpretation, the problem can be shown to be in PP by combining the argument in the proof of Theorem 12 showing how to average over consistent totally ordered plans with the argument in the proof of Theorem 3 showing how to evaluate a plan in a propositional domain in PP. Alternatively, we could express the evaluation of a partially ordered plan under the average interpretation as a compact probabilistic acyclic plan; Corollary 5 states that such plans can be evaluated in PP. PP-hardness follows directly from Theorem 3, because totally ordered plans are a special case of partially ordered plans and evaluating totally ordered plans is PP-hard. ∎

## 5. Applications

To help illustrate the utility of our results, this section cites several planners from the literature and analyzes the computational complexity of the problems they attack. We do not give detailed explanations of the planners themselves; for this, we refer the reader to the original papers. We focus on three planning systems: WITNESS (Brown University), BURIDAN (University of Washington), and TREEPLAN (University of British Columbia). In the process of making connections to these planners, we also describe how our work relates to the discounted-reward criterion, partial observability, other domain representations, partial order conditional planning, policy-based planning, and approximate planning.

### 5.1 Witness

The WITNESS algorithm (Cassandra, Kaelbling, & Littman, 1994; Kaelbling et al., 1998) solves flat partially observable Markov decision processes using a dynamic-programming approach. The basic algorithm finds optimal unrestricted solutions to finite-horizon problems. Papadimitriou and Tsitsiklis (1987) showed that the plan-existence problem for polynomial-horizon partially observable Markov decision processes is PSPACE-complete.

As an extension to their finite-horizon algorithm, Kaelbling et al. (1998) sketch a method for finding optimal looping plans for some domains. Although this is not presented as a formal algorithm, it is not unreasonable to say that the pure form of the problem that this extended version of WITNESS attacks is one of finding a valid polynomial-size looping plan for a partially observable domain. The similarities between this problem and that described in Section 3 are that the domains are flat and that the plans are identical in form. The apparent differences are that WITNESS optimizes a reward function instead of probability of goal satisfaction and that WITNESS works in partially observable domains whereas our results are defined in terms of completely observable domains. Both of these apparent differences are insignificant, however, from a computational complexity point of view.

First, WITNESS attempts to maximize the expected total discounted reward over an infinite horizon (sometimes called optimizing a time-separable value function). As argued by Condon (1992), any problem defined in terms of a sum of discounted rewards can be recast as one of goal satisfaction. The argument proceeds roughly as follows. Let $0 < \gamma < 1$ be the discount factor and $R(s, a)$ be the immediate reward received for taking action $a$ in state $s$.

Define

$$R'(s, a) = \frac{R(s, a) - \min_{s', a'} R(s', a')}{\max_{s', a'} R(s', a') - \min_{s', a'} R(s', a')}.$$

From this, we have that $0 \leq R'(s, a) \leq 1$ for all $s$ and $a$ and that the value of any plan with respect to the revised reward function is a simple linear transformation of its true value. Now, we introduce an auxiliary state $g$ to be the goal state and create a new transition function $t'$ such that $t'(s, a, g) = (1-\gamma)R'(s, a)$ and $t'(s, a, s') = (1-(1-\gamma)R'(s, a))t(s, a, s')$ for $s' \neq g$; $t'$ is a well-defined transition function and the probability of goal satisfaction for any plan under transition function $t'$ is precisely the same as the expected total discounted reward under reward function $R'$ and transition function $t$. Thus, any problem stated as one of optimizing the expected total of discounted immediate rewards can be turned into an equivalent problem of optimizing goal satisfaction with only a slight change to the transition function and one additional state. This means there is no fundamental computational complexity difference between these two different types of planning objectives.

The second apparent difference between the problem solved by the extended WITNESS algorithm and that described in Section 3 is that of partial versus complete observability. In fact, our results do address partial observability, albeit indirectly. In our formulation of the plan-existence problem, plans are constrained to make no conditional branches (in the totally ordered and partially ordered cases), or to branch only on distinctions made by the step-transition function $\delta$ (in the acyclic and looping cases); these two choices correspond to unobservable and partially observable domains, respectively. In a partially observable domain, the plan-existence problem becomes one of finding a valid polynomial-size finite-state controller subject to the given observability constraints. Nothing in our complexity proofs depends on the presence or absence of additional observability constraints. Therefore, it is a direct corollary of Theorem 2 that the plan-existence problem for polynomial-horizon plans in unobservable domains is NP-complete (Papadimitriou & Tsitsiklis, 1987) and of Theorem 7 that the plan-existence problem for polynomial-size looping plans in partially observable domains is NP-complete (this is a new result).

It is interesting to note that the computational complexity of searching for size-bounded plans in partially observable domains is generally substantially less than that of solving the corresponding unconstrained partially observable Markov decision process. For example, we found that the plan-existence problem for acyclic plans in propositional domains is NP$^{\text{PP}}$-complete (Theorem 4). The corresponding unconstrained problem is that of determining the existence of a history-dependent policy for a polynomial-horizon, compactly represented partially observable Markov decision process, which is EXPSPACE-complete (Theorem 4.15 of Goldsmith et al., 1996, or Theorem 6.8 of Mundhenk et al., 1997b). The gap here is enormous: EXPSPACE is to EXP what PSPACE is to P, and EXP is already provably intractable in the worst case. In contrast to EXPSPACE-complete problems, it is conceivable that good heuristics for NP$^{\text{PP}}$-complete problems can be created by extensions of recent advances in heuristics for NP-complete problems. Therefore, there is some hope of devising effective planning algorithms by building on the observations in this paper and searching for optimal size-bounded plans instead of optimal unrestricted plans; in fact, recent planners for both propositional domains (Majercik & Littman, 1998a, 1998b) and flat domains (Hansen, 1998) are motivated by these results.

| Domain Type | Horizon Type | **Size-Bounded Plan** | **Unrestricted Plan** |
|---|---|---|---|
| flat | polynomial | NP-complete | PSPACE-complete |
| propositional | polynomial | NP$^{\text{PP}}$-complete | EXPSPACE-complete |
| flat | infinite | NP-complete | undecidable |
| propositional | infinite | PSPACE-complete | undecidable |

Table 3: Complexity results for plan existence in partially observable domains

Table 3 summarizes complexity results for planning in partially observable domains. The results for size-bounded plans are corollaries of Theorems 2, 4, 7, and 9 of this paper. The results for unrestricted plans are due to Papadimitriou and Tsitsiklis (1987) (flat, polynomial), Goldsmith et al. (1996) (propositional, polynomial), and Hanks (1996) (infinite-horizon). This last result is derived by noting the isomorphism of the infinite-horizon problem to the emptiness problem for probabilistic finite-state automata, which is undecidable (Rabin, 1963).

## 5.2 Buridan

The BURIDAN planner (Kushmerick et al., 1995) finds partially ordered plans for propositional domains in the PSO representation. There are two identifiable differences between the problem solved by BURIDAN and the problem analyzed in Section 4: the representation of planning problems and the fact that BURIDAN is not restricted to find polynomial-size plans. We address each of these differences below.

Although, on the surface, PSO is different from ST, either can be converted into the other in polynomial time with at most a polynomial increase in domain size. In particular, the effect of an action in PSO is represented by a single decision tree consisting of proposition nodes (like ST) and random nodes (easily simulated in ST using auxiliary propositions). At the leaves are a list of propositions that become true and another list of propositions that become false should that leaf be reached. This type of correlated effect is also easily represented in ST using the chain rule of probability theory to decompose the probability distribution into separate probabilities for each proposition and careful use of the ":**new**" suffix. Thus, any PSO domain can be converted to a similar size ST domain quickly.

Similarly, a domain in ST can be converted to PSO with at most a polynomial expansion. This conversion is too complex to sketch here, but follows from the proof of equivalence between ST and a simplified representation called IF (Littman, 1997a). Given the polynomial equivalence between ST and PSO, any complexity results for ST carry over to PSO.[5]

The results described in this paper concern planning problems in which a bound is given on the size of the plan sought. Although Kushmerick et al. (1995) do not explicitly describe their planner as one that prefers small plans to large plans, the design of the planner as one that searches through the space of plans makes the notion of plan size central to the algorithm. Indeed, the public-domain BURIDAN implementation uses plan size as part of a best-first search procedure for identifying a sufficiently successful plan. This means that, all other things being equal, shorter plans will be found before larger plans. Furthermore, to assure termination, the planner only considers a fixed number of plans before halting, thus putting a limit indirectly on the maximum allowable plan size. So, although BURIDAN does not attempt to solve precisely the same problem that we considered, it is fair to say that the problem we consider is an idealization of the problem attacked by BURIDAN. Regardless, our lower bounds on complexity apply to BURIDAN.

---

5. To be more precise, this is true for complexity classes closed under log-space reductions.

Kushmerick et al. (1995) looked at generating sufficiently successful plans under both the optimistic interpretation and the pessimistic interpretation. They also explicitly examined the plan-evaluation problem for partially ordered plans under both interpretations. Therefore, Theorems 13 and 14 apply to BURIDAN.

The more sophisticated C-BURIDAN planner (Draper et al., 1994) extends BURIDAN to plan in partially observable domains and to produce plans with conditional execution. The results of our work also shed light on the computational complexity of the problem addressed by C-BURIDAN. Draper et al. (1994) devised a representation for partially ordered acyclic (conditional) plans. In this representation, each plan step generates an *observation label* as a function of the probabilistic outcome of the step. Each step also has an associated set of *context labels* dictating the circumstances under which that step must be executed. A plan step is executed only if its context labels are consistent with the observation labels produced in earlier steps. In its totally ordered form, this type of plan can be expressed as a compact acyclic plan; Corollary 5 can be used to show that the plan-evaluation and plan-existence problems for a totally ordered version of C-BURIDAN's conditional plan representation in propositional domains are PP-complete and $NP^{PP}$-complete, respectively.

In our results above, we consider evaluating and searching for plans that are partially ordered and plans that have conditional execution, but not both at once. Nonetheless, the same sorts of techniques presented in this paper can be applied to analyzing the problems attacked by C-BURIDAN. For example, consider the plan-existence problem for C-BURIDAN's partially ordered conditional plans under the optimistic interpretation. This problem asks whether there is a partially ordered conditional plan that has some total order that reaches the goal with sufficient probability. This is equivalent to asking whether there is a totally ordered conditional plan that reaches the goal with sufficient probability. Therefore, the problem is $NP^{PP}$-complete, by the argument in the previous paragraph.

In spite of many superficial differences between the problems analyzed in this paper and those studied by the creators of the BURIDAN planners, our results are quite relevant to understanding their work.

## 5.3 Treeplan

A family of planners have been designed that generate a decision-tree-based representation of stationary policies (mappings from state to action) (Boutilier et al., 1995; Boutilier & Poole, 1996; Boutilier & Dearden, 1996) in probabilistic propositional domains; we refer to these planners collectively as the TREEPLAN planners. Once again, these planners solve problems that are not identical to the problems addressed in this paper but are closely related.

The planner described by Boutilier et al. (1995) finds solutions that maximize expected total discounted reward in compactly represented Markov decision processes (the domain representation used is expressively equivalent to ST). As mentioned earlier, the difference between maximizing goal satisfaction and maximizing expected total discounted reward is a superficial one, so the problem addressed by this planner is EXP-complete (Littman, 1997a). Although the policies used by Boutilier et al. (1995) appears quite dissimilar from the finite-state controllers described in our work, policies can be converted to a type of similarly sized compact *looping* plan (an extension of the type of plan described in Corollary 5). The conversion from stationary policies to looping plans is as described in the proof of Theorem 7, except that the resulting plans are represented compactly.

In later work, Boutilier and Dearden (1996) show how it is possible to limit the size of the representation of the policy in TREEPLAN and still obtain approximately optimal performance. This is necessary because, in general, the size of decision trees needed to represent the optimal policies can be exponentially large. By keeping the decision trees from getting too large, the resulting planner becomes subject to an extension of Theorem 9 and, therefore, attacks a PSPACE-complete problem.

One emphasis of Boutilier and Dearden (1996) is on finding approximately optimal solutions, with the hope that doing so is easier than finding optimal solutions. We do not explore the worst-case complexity of approximation in this paper, although Lusena, Goldsmith, and Mundhenk (1998) have produced some strong negative results in this area. A related issue is one of using simulation (random sampling) to find approximately optimal solutions to probabilistic planning problems. Some empirical successes have been obtained with the related approach of reinforcement learning (Tesauro, 1994; Crites & Barto, 1996), but, once again, the worst-case complexity of probabilistic planning is not known to be any lower for approximation by simulation.

## 6. Conclusions

In this paper, we explored the computational complexity of plan evaluation and plan existence in probabilistic domains. We found that, in compactly represented propositional domains, restricting the size and form of the policies under consideration reduced the computational complexity of plan existence from EXP-complete for unrestricted plans to PSPACE-complete for polynomial-size looping plans and $\mathrm{NP}^{\mathrm{PP}}$-complete for polynomial-size acyclic plans. In contrast, in flat domains, restricting the form of the policies under consideration *increased* the computational complexity of plan existence from P-complete for unrestricted plans to NP-complete for totally ordered plans; this is because a plan that is smaller than the domain in which it operates is often unable to exploit important Markov properties of the domain. We were able to characterize precisely the complexity of all problems we examined with regard to the current state of knowledge in complexity theory.

Several problems we studied turned out to be $\mathrm{NP}^{\mathrm{PP}}$-complete. The class $\mathrm{NP}^{\mathrm{PP}}$ promises to be very useful to researchers in uncertainty in artificial intelligence because it captures the type of problems resulting from choosing ("guessing") a solution and then evaluating its probabilistic behavior. This is precisely the type of problem faced by planning algorithms in probabilistic domains, and captures important problems in other domains as well, such as constructing explanations in belief networks and designing robust communication networks. We provide a new conceptually simple $\mathrm{NP}^{\mathrm{PP}}$-complete problem, E-MAJSAT, that may be useful in further explorations in this direction.

The basic structure of our results is that if plan evaluation is complete for some class $\mathcal{C}$, then plan existence is typically $\mathrm{NP}^{\mathcal{C}}$-complete. This same basic structure holds in determin-istic domains: evaluating a totally ordered plan in a propositional domain is P-complete (for sufficiently powerful domain representations) and determining the existence of a polynomial-size totally ordered plan is $NP^P = NP$-complete.

From a pragmatic standpoint, the intuition that searching for small plans is more efficient than searching for arbitrary size plans suggests that exact dynamic-programming algorithms, which are so successful in flat domains, may not be as effective in propositional domains; they do not focus their efforts on the set of small plans. Algorithm-development energy, therefore, might fruitfully be spent devising heuristics for problems in the class $NP^{PP}$ as this class captures the essence of searching for small plans in probabilistic domains—some early results in this direction are appearing (Majercik & Littman, 1998a, 1998b). Complexity theorists have only recently begun to explore classes such as $NP^{PP}$ that lie between the polynomial hierarchy and PSPACE and algorithm designers have come to these classes even more recently. As this paper marks the beginning of our exploration of this class of problems, much work is still to be done in probing algorithmic implications, but it is our hope that heuristics for $NP^{PP}$ could lead to powerful methods for solving a range of important uncertainty-sensitive combinatorial problems.

## Acknowledgements

This work was supported in part by grants NSF-IRI-97-02576-CAREER (Littman), and NSF CCR-9315354 (Goldsmith). We gratefully acknowledge Andrew Klapper, Anne Condon, Matthew Levy, Steve Majercik, Chris Lusena, Mark Peot, and our reviewers for helpful feedback and conversations on this topic.

## Appendix A. Complexity of E-MAJSAT

The E-MAJSAT problem is: given a pair $(\phi, k)$ consisting of a Boolean formula $\phi$ of $n$ variables $x_1, \ldots, x_n$ and a number $1 \leq k \leq n$, is there an assignment to the first $k$ variables $x_1, \ldots, x_k$ such that the majority of assignments to the remaining $n-k$ variables $x_{k+1}, \ldots, x_n$ satisfies $\phi$?

For $k = n$, this is precisely Boolean satisfiability, a classic NP-complete problem. This is because we are asking whether there exists an assignment to *all* the variables that makes $\phi$ true. For $k = 0$, E-MAJSAT is precisely MAJSAT, a well-known PP-complete problem. This is because we are asking whether the majority of all *total* assignments makes $\phi$ true.

Deciding an instance of E-MAJSAT for intermediate values of $k$ has a different character. It involves both an NP-type calculation to pick a good setting for the first $k$ variables and a PP-type calculation to see if the majority of assignments to the remaining variables makes $\phi$ true. This is akin to searching for a good answer (plan, schedule, coloring, belief network explanation, etc.) in a combinatorial space when "good" is determined by a computation over probabilistic quantities. This is just the type of computation described by the class $NP^{PP}$, and we show next that E-MAJSAT is $NP^{PP}$-complete.

**Theorem 15** E-MAJSAT *is* $NP^{PP}$-*complete.*

**Proof:** Membership in $\text{NP}^{\text{PP}}$ follows directly from definitions. To show completeness of E-MAJSAT, we first observe (Torán, 1991) that $\text{NP}^{\text{PP}}$ is the $\leq_m^{\text{NP}}$-closure of the PP-complete set MAJSAT. Thus, any $\text{NP}^{\text{PP}}$ computation can be modeled by a nondeterministic machine $N$ that, on each possible computation, first guesses a sequence $s$ of bits that controls its nondeterministic moves, deterministically performs some computation on input $x$ and $s$, and then writes down a formula $q_{x,s}$ with variables in $z_1, \ldots, z_l$ as a query to MAJSAT. Finally, $N(x)$ with oracle MAJSAT accepts if and only if for some $s$, $q_{x,s} \in$ MAJSAT.

Given any input $x$, like in Cook's Theorem, we can construct a formula $\phi_x$ with variables $y_1, \ldots, y_k$ and $z_1, \ldots, z_l$ such that for every assignment $a_1, \ldots, a_k, b_1, \ldots, b_l$ it holds that $\phi_x(a_1, \ldots, a_k, b_1, \ldots, b_l) = q_{x, a_1 \cdots a_k}(b_1, \ldots, b_l)$. Thus, $(\phi_x, k) \in$ E-MAJSAT if and only if for some assignment $s$ to $y_1, \ldots, y_k$, $q_{x,s} \in$ MAJSAT if and only if $N(x)$ accepts. ∎

# References

Allender, E., & Ogihara, M. (1996). Relationships among PL, #L, and the determinant. *Theoretical Informatics and Applications, 30*(1), 1–21.

Àlvarez, C., & Jenner, B. (1993). A very hard log-space counting class. *Theoretical Computer Science, 107*, 3–30.

Bäckström, C. (1995). Expressive equivalence of planning formalisms. *Artificial Intelligence, 76*(1–2), 17–34.

Bäckström, C., & Nebel, B. (1995). Complexity results for SAS+ planning. *Computational Intelligence, 11*(4), 625–655.

Balcázar, J., Díaz, J., & Gabarró, J. (1988/1990). *Structural Complexity I/II.* EATCS Monographs on Theoretical Computer Science. Springer Verlag.

Borodin, A., Cook, S., & Pippenger, N. (1983). Parallel computation for well-endowed rings and space-bounded probabilistic machines. *Information and Control, 58*(1–3), 113–136.

Boutilier, C., Dean, T., & Hanks, S. (1998). Decision theoretic planning: Structural assumptions and computational leverage. In preparation.

Boutilier, C., & Dearden, R. (1996). Approximating value trees in structured dynamic programming. In Saitta, L. (Ed.), *Proceedings of the Thirteenth International Conference on Machine Learning.*

Boutilier, C., Dearden, R., & Goldszmidt, M. (1995). Exploiting structure in policy construction. In *Proceedings of the Fourteenth International Joint Conference on Artificial Intelligence*, pp. 1104–1113.

Boutilier, C., & Poole, D. (1996). Computing optimal policies for partially observable decision processes using compact representations. In *Proceedings of the Thirteenth National Conference on Artificial Intelligence*, pp. 1168–1175. AAAI Press/The MIT Press.

Bylander, T. (1994). The computational complexity of propositional STRIPS planning. *Artificial Intelligence*, *69*, 161–204.

Cassandra, A. R., Kaelbling, L. P., & Littman, M. L. (1994). Acting optimally in partially observable stochastic domains. In *Proceedings of the Twelfth National Conference on Artificial Intelligence*, pp. 1023–1028 Seattle, WA.

Chapman, D. (1987). Planning for conjunctive goals. *Artificial Intelligence*, *32*, 333–379.

Condon, A. (1992). The complexity of stochastic games. *Information and Computation*, *96*(2), 203–224.

Crites, R. H., & Barto, A. G. (1996). Improving elevator performance using reinforcement learning. In Touretzky, D. S., Mozer, M. C., & Hasselmo, M. E. (Eds.), *Advances in Neural Information Processing Systems 8* Cambridge, MA. The MIT Press.

Dearden, R., & Boutilier, C. (1997). Abstraction and approximate decision-theoretic planning. *Artificial Intelligence*, *89*(1–2), 219–283.

Draper, D., Hanks, S., & Weld, D. (1994). Probabilistic planning with information gathering and contingent execution. In *Proceedings of the AAAI Spring Symposium on Decision Theoretic Planning*, pp. 76–82.

Drummond, M., & Bresina, J. (1990). Anytime synthetic projection: Maximizing the probability of goal satisfaction. In *Proceedings of the Eighth National Conference on Artificial Intelligence*, pp. 138–144. Morgan Kaufmann.

Erol, K., Nau, D. S., & Subrahmanian, V. S. (1995). Complexity, decidability and undecidability results for domain-independent planning. *Artificial Intelligence*, *76*, 75–88.

Gill, J. (1977). Computational complexity of probabilistic Turing machines. *SIAM Journal on Computing*, *6*(4), 675–695.

Goldman, R. P., & Boddy, M. S. (1994). Epsilon-safe planning. In *Proceedings of the 10th Conference on Uncertainty in Artificial Intelligence (UAI94)*, pp. 253–261 Seattle, WA.

Goldsmith, J., Littman, M., & Mundhenk, M. (1997a). The complexity of plan existence and evaluation in probabilistic domains. Tech. rep. CS-1997-07, Department of Computer Science, Duke University.

Goldsmith, J., Littman, M. L., & Mundhenk, M. (1997b). The complexity of plan existence and evaluation in probabilistic domains. In *Proceedings of the Thirteenth Annual Conference on Uncertainty in Artificial Intelligence (UAI–97)*, pp. 182–189 San Francisco, CA. Morgan Kaufmann Publishers.

Goldsmith, J., Lusena, C., & Mundhenk, M. (1996). The complexity of deterministically observable finite-horizon Markov decision processes. Tech. rep. 268-96, Department of Computer Science, University of Kentucky.

Hanks, S. (1996). Decision-theoretic planning in unobservable domains is undecidable. Personal communication.

Hansen, E. A. (1998). *Finite-Memory Control of Partially Observable Systems*. Ph.D. thesis, University of Massachusetts.

Jung, H. (1985). On probabilistic time and space. In *Proceedings 12th ICALP*, pp. 281–291. Lecture Notes in Computer Science, Springer-Verlag.

Kaelbling, L. P., Littman, M. L., & Cassandra, A. R. (1998). Planning and acting in partially observable stochastic domains. *Artificial Intelligence*, *101*(1–2), 99–134.

Koenig, S., & Simmons, R. G. (1994). Risk-sensitive planning with probabilistic decision graphs. In *Proceedings of the 4th International Conference on Principles of Knowledge Representation and Reasoning*, pp. 363–373.

Kushmerick, N., Hanks, S., & Weld, D. S. (1995). An algorithm for probabilistic planning. *Artificial Intelligence*, *76*(1-2), 239–286.

Ladner, R. (1989). Polynomial space counting problems. *SIAM Journal on Computing*, *18*, 1087–1097.

Lin, S.-H., & Dean, T. (1995). Generating optimal policies for high-level plans with conditional branches and loops. In *Proceedings of the Third European Workshop on Planning*, pp. 205–218.

Littman, M. L. (1997a). Probabilistic propositional planning: Representations and complexity. In *Proceedings of the Fourteenth National Conference on Artificial Intelligence*, pp. 748–754. AAAI Press/The MIT Press.

Littman, M. L. (1997b). Solving large POMDPs: Lessons from complexity theory. Talk presented at the DARPA AI Workshop in Providence, RI. Slides available at URL http://www.cs.duke.edu/~mlittman/talks/darpa97-pomdp.ps.

Lovejoy, W. S. (1991). A survey of algorithmic methods for partially observable Markov decision processes. *Annals of Operations Research*, *28*(1), 47–65.

Lusena, C., Goldsmith, J., & Mundhenk, M. (1998). Nonapproximability results for Markov decision processes. Tech. rep. UK CS Dept TR 275-98, University of Kentucky.

Majercik, S. M., & Littman, M. L. (1998a). MAXPLAN: A new approach to probabilistic planning. In Simmons, R., Veloso, M., & Smith, S. (Eds.), *Proceedings of the Fourth International Conference on Artificial Intelligence Planning*, pp. 86–93. AAAI Press.

Majercik, S. M., & Littman, M. L. (1998b). Using caching to solve larger probabilistic planning problems. In *Proceedings of the Fifteenth National Conference on Artificial Intelligence*, pp. 954–959. The AAAI Press/The MIT Press.

Mansell, T. M. (1993). A method for planning given uncertain and incomplete information. In *Proceedings of the 9th Conference on Uncertainty in Artificial Intelligence*, pp. 350–358. Morgan Kaufmann Publishers.

McAllester, D., & Rosenblitt, D. (1991). Systematic nonlinear planning. In *Proceedings of the 9th National Conference on Artificial Intelligence*, pp. 634–639.

Mundhenk, M., Goldsmith, J., & Allender, E. (1997a). The complexity of policy-evaluation for finite-horizon partially-observable Markov decision processes. In *Proceedings of 22nd Symposium on Mathematical Foundations of Computer Science (published in Lecture Notes in Computer Science)*. Springer-Verlag.

Mundhenk, M., Goldsmith, J., Lusena, C., & Allender, E. (1997b). Encyclopaedia of complexity results for finite-horizon Markov decision process problems. Tech. rep. UK CS Dept TR 273-97, University of Kentucky.

Papadimitriou, C. H. (1994). *Computational Complexity*. Addison-Wesley, Reading, MA.

Papadimitriou, C. H., & Tsitsiklis, J. N. (1987). The complexity of Markov decision processes. *Mathematics of Operations Research*, *12*(3), 441–450.

Platzman, L. K. (1981). A feasible computational approach to infinite-horizon partially-observed Markov decision problems. Tech. rep. J-81-2, Georgia Institute of Technology, Atlanta, GA.

Puterman, M. L. (1994). *Markov Decision Processes—Discrete Stochastic Dynamic Programming*. John Wiley & Sons, Inc., New York, NY.

Rabin, M. O. (1963). Probabilistic automata. *Information and Control*, *6*(3), 230–245.

Roth, D. (1996). On the hardness of approximate reasoning. *Artificial Intelligence*, *82*(1–2), 273–302.

Simon, J. (1975). *On some central problems in computational complexity*. Ph.D. thesis, Cornell University. Also Cornell Department of Computer Science Technical Report TR75-224.

Smith, D. E., & Williamson, M. (1995). Representation and evaluation of plans with loops. Working notes for the 1995 Stanford Spring Symposium on Extended Theories of Action.

Tesauro, G. (1994). TD-Gammon, a self-teaching backgammon program, achieves master-level play. *Neural Computation*, *6*(2), 215–219.

Toda, S. (1991). PP is as hard as the polynomial-time hierarchy. *SIAM Journal on Computing*, *20*, 865–877.

Torán, J. (1991). Complexity classes defined by counting quantifiers. *Journal of the ACM*, *38*(3), 753–774.

Vinay, V. (1991). Counting auxiliary pushdown automata and semi-unbounded arithmetic circuits. In *Proc. 6th Structure in Complexity Theory Conference*, pp. 270–284. IEEE.